\documentclass{article}
\usepackage{iclr2026_conference,times}
\setcitestyle{numbers,square,citesep={,}}

\usepackage{amsmath,amsfonts,bm}

\def\eqref#1{equation~\ref{#1}}

\def\1{\bm{1}}

\DeclareMathAlphabet{\mathsfit}{\encodingdefault}{\sfdefault}{m}{sl}
\SetMathAlphabet{\mathsfit}{bold}{\encodingdefault}{\sfdefault}{bx}{n}

\usepackage{hyperref}
\usepackage{url}
\usepackage{graphicx}
\usepackage{booktabs}
\usepackage{threeparttable}
\usepackage{multirow}
\usepackage{array}
\usepackage[table]{xcolor}
\usepackage{xspace}
\usepackage{enumitem}
\usepackage{arydshln}
\usepackage{pifont}
\IfFileExists{fontawesome5.sty}{%
  \usepackage{fontawesome5}%
  \newcommand{\codeicon}{\faIcon{github}}%
  \newcommand{\webicon}{\faIcon{globe}}%
}{%
  \newcommand{\codeicon}{\texttt{[code]}}%
  \newcommand{\webicon}{\texttt{[web]}}%
}
\usepackage{xcolor}
\usepackage{listings}
\newcommand{\cmark}{\textcolor{green!70!black}{\ding{51}}}
\newcommand{\xmark}{\textcolor{red}{\ding{55}}}

\usepackage{adjustbox}

\definecolor{trackgray}{gray}{0.92}
\usepackage{tabularx}
\usepackage{makecell}
\usepackage{ragged2e}

\newcommand{\Method}{\textsc{AutoMedBench}\xspace}
\newcommand{\totaltasknum}{48}

\newcommand{\titleucsclogoheight}{0.62in}
\newcommand{\titlenvidialogoheight}{0.50in}
\definecolor{bestgreen}{RGB}{220,245,220}

\newcommand{\titlelogos}{%
  \makebox[\textwidth]{%
    \makebox[0.48\textwidth][l]{%
      \includegraphics[height=\titleucsclogoheight]{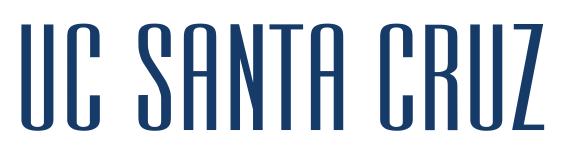}%
    }%
    \hfill
    \makebox[0.48\textwidth][r]{%
      \includegraphics[height=\titlenvidialogoheight]{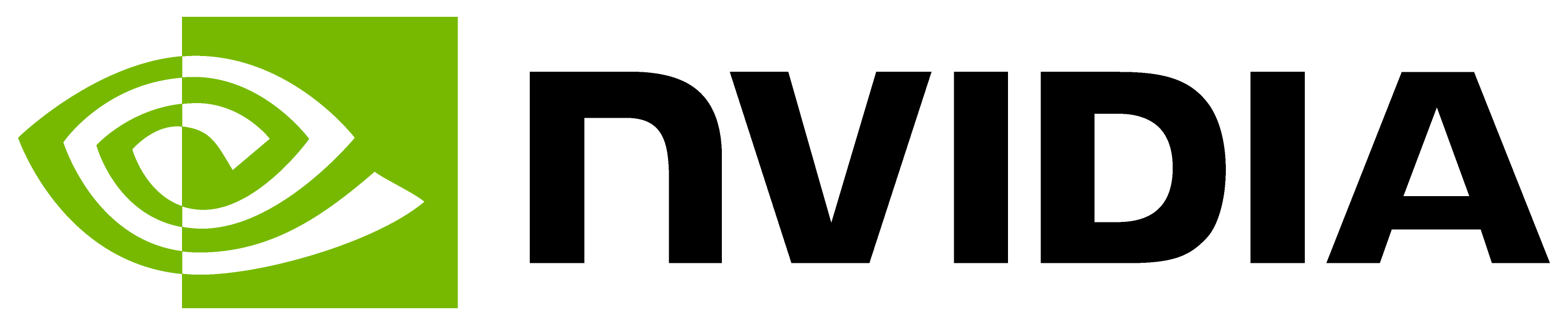}%
    }%
  }%
}

\fancypagestyle{automedtitle}{%
  \fancyhf{}%
  \fancyfoot[C]{\thepage}%
  \renewcommand{\headrulewidth}{0pt}%
}

\newcommand{\placetitlelogos}{%
  \AddToShipoutPicture*{%
    \AtPageUpperLeft{%
      \put(\LenToUnit{1.5in},\LenToUnit{-0.66in}){%
        \titlelogos
      }%
      \put(\LenToUnit{1.5in},\LenToUnit{-0.80in}){%
        \rule{\textwidth}{0.4pt}%
      }%
    }%
  }%
}

\lstdefinestyle{appendixtrace}{
  basicstyle=\ttfamily\scriptsize,
  breaklines=true,
  breakatwhitespace=false,
  breakindent=0pt,
  postbreak={},
  columns=fullflexible,
  keepspaces=true,
  showstringspaces=false,
  frame=single,
  framerule=0.2pt,
  aboveskip=6pt,
  belowskip=6pt,
  xleftmargin=0pt,
  xrightmargin=0pt
}

\title{\Method: \\ Towards Medical AutoResearch with Agentic AI Models}

\author{
Junqi Liu$^{1}$ \quad
Selena Song$^{1}$ \quad
Yuhan Wang$^{1}$ \quad
Jiawei Mao$^{1}$ \quad
Hardy Chen$^{1}$\\
Xiaoke Huang$^{1}$ \quad
Tianhao Qi$^{1}$ \quad
Pengfei Guo$^{2}$ \quad
Yucheng Tang$^{2}$ \quad
Yufan He$^{2}$ \\
Can Zhao$^{2}$ \quad
Andriy Myronenko$^{2}$ \quad
Dong Yang$^{2}$ \quad
Daguang Xu$^{2}$ \quad
Yuyin Zhou$^{1}$\\
\\
$^{1}$University of California, Santa Cruz \quad
$^{2}$NVIDIA\\[0.45em]
{\normalfont\normalsize
\codeicon\ GitHub: \href{https://github.com/AutoMedBench/AutoMedBench}{https://github.com/AutoMedBench/AutoMedBench}}\\[0.2em]
{\normalfont\normalsize
\webicon\ Leaderboard: \href{https://automedbench.github.io}{https://automedbench.github.io}}
}

\iclrfinalcopy

\begin{document}
\placetitlelogos
\maketitle
\thispagestyle{automedtitle}

\begin{abstract}
Autonomous agents are increasingly expected to support end-to-end medical-AI research workflows, moving beyond isolated prediction tasks or short-form clinical question answering.
However, existing medical agent benchmarks primarily evaluate final outputs, providing limited visibility into agent behavior within the research process.
In long-horizon workflows, this final-output view becomes insufficient: agents must preserve context across many interaction turns, while failures can emerge from different workflow stages and compound before being collapsed into a single end score.
To address this gap, we present \Method, a workflow-aware benchmark for autonomous medical-AI research across diverse medical imaging and multimodal inference tasks. 
\Method organizes agent execution into a unified five-stage workflow (S1--S5): \emph{Plan}, \emph{Setup}, \emph{Validate}, \emph{Inference}, and \emph{Submit}. 
It comprises long-horizon tasks with each run averaging 33 agent turns, spanning five research tracks: segmentation, image enhancement, visual question answering (VQA), report generation, and lesion detection. 
Each task is evaluated under two difficulty tiers, \textsc{Lite} and \textsc{Standard}, which use the same data and metrics but differ in the amount of task-brief scaffolding. 
Each run is scored using both final task performance and S1--S5 stage scores, enabling stage-level analysis from the initial task brief to the final submitted artifact.
Across thousands of recorded runs, stage-level scoring reveals that \emph{Validate} is the weakest workflow stage on average, whereas \emph{Setup} is the strongest, suggesting that current agents are better at making pipelines executable than at verifying their reliability.
Post-run error analysis further shows that verification and submission failures dominate the tagged errors, accounting for 37.7\% and 38.1\% of fired codes, respectively, whereas task-understanding errors are rare at 0.9\%.
These error codes are not merely descriptive: runs with one fired error code have a 48\% lower overall score than runs with no error code on average.
By linking stage-level scores with diagnostic error codes, \Method exposes hidden breakdowns, including failed model loading, shape bugs, skipped validation, empty outputs, and malformed submissions, that are often obscured by final-output metrics alone.
These findings suggest that strong medical research agents must combine high-quality domain knowledge with robust engineering capabilities, including intermediate validation and error recovery throughout the workflow.
\end{abstract}

\section{Introduction}
\begin{figure}[htbp]
    \centering
    \includegraphics[width=\linewidth]{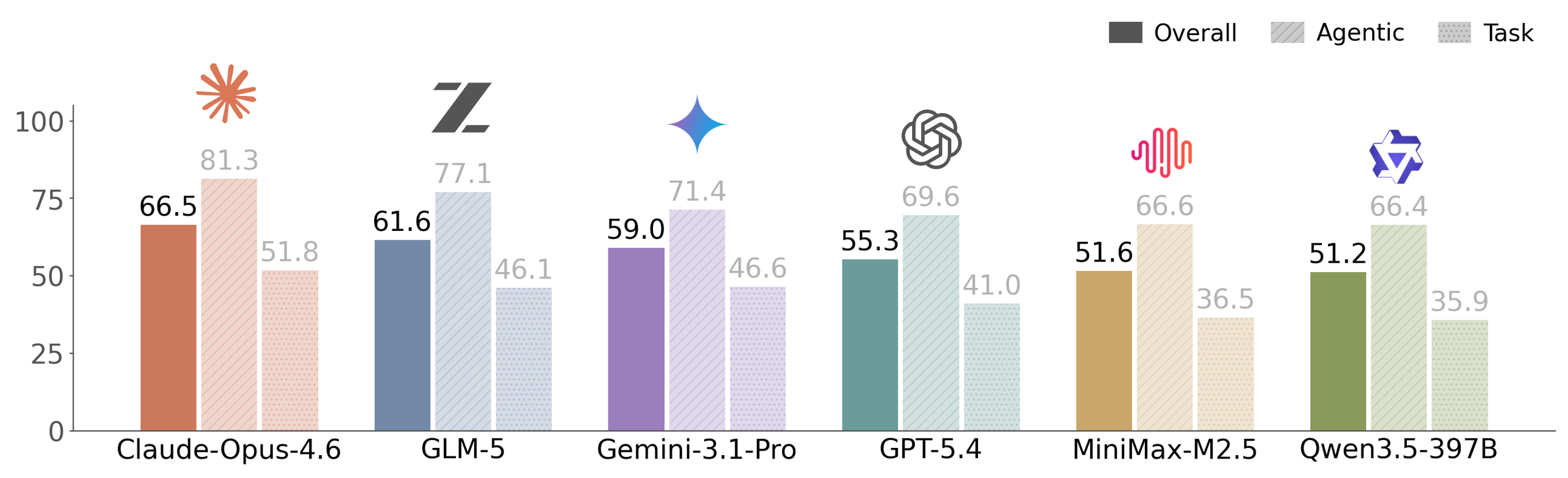}
    \caption{\textbf{Overall leaderboard.} Overall, agentic, and task scores for the 6 evaluated agents. Agents are ranked by overall scores. The overall score averages the workflow-based agentic score and the held-out task score. Per-track leaderboards are in \figureautorefname~\ref{fig:task_wise_leaderboard}.}
    \label{fig:leaderboard}
\end{figure}

Large language model agents are rapidly moving beyond passive question answering toward autonomous research assistance. Equipped with code execution, tool use, long-context reasoning, and access to external resources, these systems are increasingly expected to plan experiments, configure environments, run pipelines, inspect intermediate results, and produce research artifacts~\citep{agentbench2024,swebench2023,mlebench2024,paperbench2025,gao2026camyla}. This emerging direction, often referred to as AutoResearch, raises a fundamental question: can current agentic AI systems conduct research workflows that would otherwise require human researchers?

Medical AI provides a particularly important and challenging testbed for this question. Unlike many single-step reasoning or coding tasks, medical-AI research requires agents to combine domain understanding with robust engineering execution. A typical task may require interpreting a clinical or biomedical objective, handling heterogeneous imaging modalities, selecting an appropriate model or algorithm, resolving dependencies, validating intermediate outputs, running inference at scale, and submitting artifacts in a strict evaluation format~\citep{litjens2017survey,isensee2021nnunet,lin2023medvqa,chen2020r2gen}. These requirements make medical AI a natural stress test for autonomous research agents: success requires not only medical knowledge, but also the ability to execute and verify a complete research workflow.

However, existing medical and healthcare agent benchmarks provide limited visibility into this setting. Many benchmarks focus on medical question answering~\citep{medqa2021,pubmedqa2019,medmcqa2022,multimedqa2023}, clinical dialogue or health scenarios~\citep{healthbench2025,medhelm2026,schmidgall2024agentclinic,hicks2026healthbenchprofessionalevaluatinglarge}, EHR/FHIR interaction~\citep{jiang2025medagentbenchrealisticvirtualehr,shi2024ehragent,lee2025fhir,liao2026agentehr,liu2026physicianbench}, healthcare administration~\citep{bedi2026healthadminbench}, or final task success. While these settings are valuable, they do not directly evaluate whether an agent can complete an end-to-end medical-AI research workflow or evaluation pipelines. More importantly, final-output metrics alone cannot reveal why an agent fails. A low score may result from misunderstanding the task, selecting an unsuitable method, failing to configure the environment, neglecting validation, producing malformed outputs, or submitting artifacts in the wrong schema. Without stage-level evaluation, it remains unclear whether current agents are limited primarily by domain knowledge, engineering reliability, verification ability, or workflow discipline. However, stage-level evaluation is not a substitute for outcome evaluation: a high stage-level agentic score does not necessarily guarantee clinically useful outputs or high accuracy.

To address this gap, we present \Method, a workflow-aware benchmark for evaluating autonomous agents on end-to-end medical-AI research tasks. \Method organizes each agent run into a unified five-stage workflow (S1--S5): \emph{Plan}, \emph{Setup}, \emph{Validate}, \emph{Inference}, and \emph{Submit}.
This design reflects the structure of practical medical-AI research workflows, where an agent must first understand the task, configure the environment and required resources, verify intermediate outputs, run full inference, and finally submit artifacts in the required format.
The benchmark covers 24 tasks across five representative medical imaging and multimodal research tracks, including segmentation, image enhancement, visual question answering, report generation, and lesion detection, spanning diverse imaging modalities such as CT, MRI, X-ray, pathology, microscopy, dental imaging, and medical video.
These tasks are long-horizon, with each run averaging 33 agent turns, requiring agents to preserve context and make consistent decisions across multiple stages.
Each task is instantiated under two difficulty tiers, \textsc{Lite} and \textsc{Standard}, which hold the underlying data, metrics, references, and submission schemas fixed while varying the amount of scaffolding provided in the task brief.
A key feature of \Method is that it evaluates both the research process and the final artifact. Each run receives an \textsc{Agentic} score based on S1--S5 workflow completion and a \textsc{Task} score based on deterministic held-out evaluation against private references. This design allows agents to be compared not only by final performance, but also by their ability to make progress through the research workflow. In addition, \Method records full interaction traces and assigns post-run cause-based error codes, enabling diagnostic analysis of where and why agent runs fail.
By linking stage-level scores with diagnostic error codes, \Method makes it possible to identify hidden workflow breakdowns that final-output metrics alone often obscure.
Our experiments with frontier base models reveal gaps between current agents and reliable autonomous medical-AI researchers. 
Across thousands of recorded runs, agents are often able to set up runnable pipelines, but validation is consistently the weakest workflow stage, indicating that they are less capable of verifying whether a pipeline is correct and reliable before scaling to full inference.
Post-run error diagnosis further assigns fired error codes across five cause-based patterns: task understanding, data or model setup, verification and recovery, implementation and execution, and deliverable submission. Verification errors, such as skipped sanity checks and ignored bad outputs, and submission errors, such as missing files and incorrect filenames, are the most frequent tagged failures, accounting for 37.7\% and 38.1\% of all fired codes, respectively, whereas task-understanding errors are rare at only 0.9\%. These error codes are also strongly associated with degraded performance: runs with one fired error code have a 48\% lower overall score than runs with no error code on average. By linking stage-level scores with diagnostic error codes, \Method exposes hidden breakdowns such as failed model loading, shape bugs, skipped validation, empty outputs, and malformed submissions, which are often obscured by final-output metrics alone. These findings suggest that the main bottleneck for current medical AutoResearch agents is not only domain knowledge, but also robust engineering execution, intermediate validation, and recovery from workflow errors.

Our contributions are threefold. First, we introduce \Method, a benchmark for evaluating autonomous medical-AI research across heterogeneous imaging and multimodal tasks using publicly available challenges and datasets, and the process of task-wise building. Second, we propose a workflow-aware evaluation protocol that combines process-level scoring and rubrics, deterministic held-out task evaluation, controlled difficulty tiers, and post-run error diagnosis. Third, we benchmark frontier LLMs, revealing the workflow stages and failure modes. Finally, we release the full execution harness and sandbox as open-source infrastructure, including containerized agents and evaluation environments with isolation.

\begin{figure}[t]
\centering
\includegraphics[width=\linewidth]{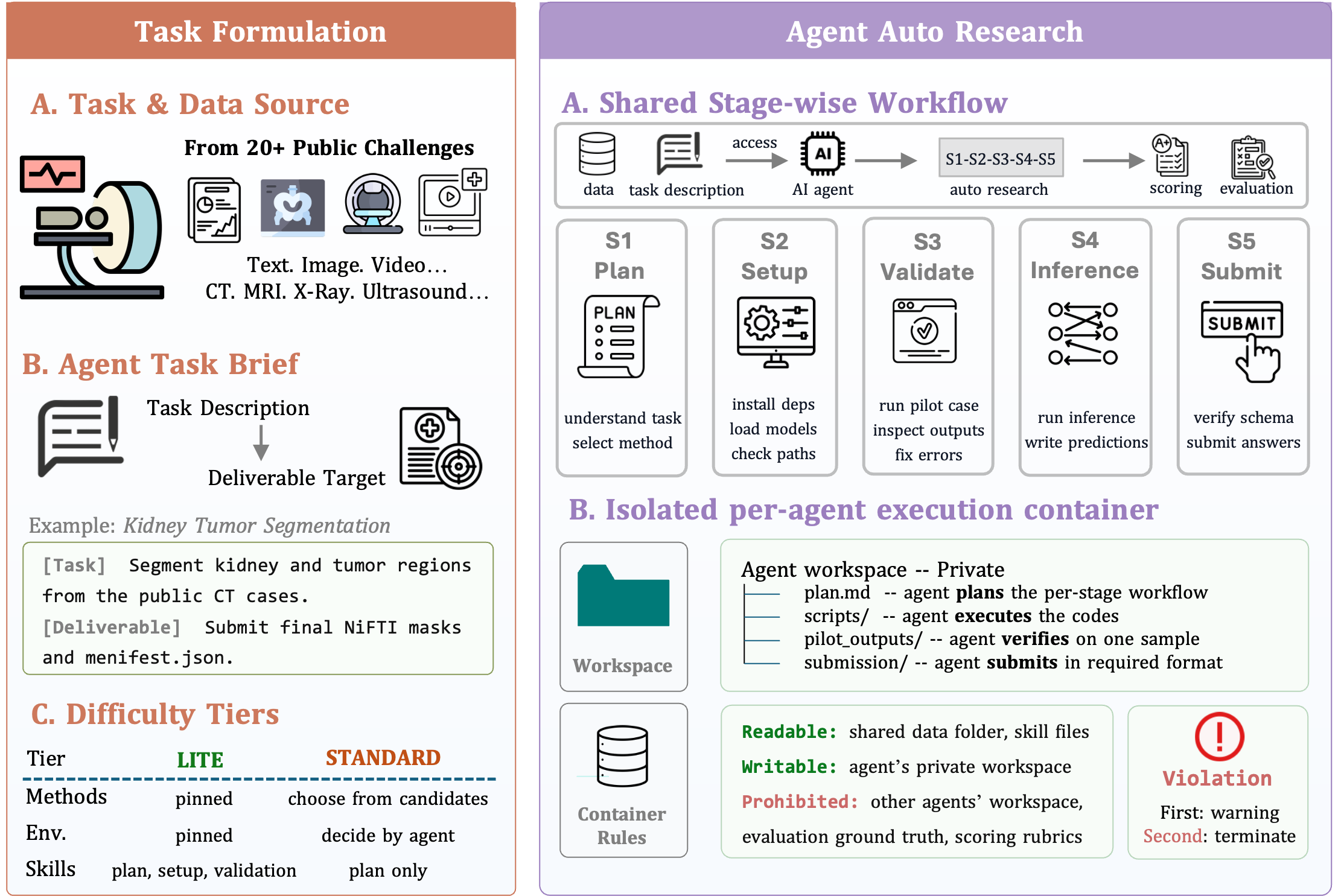}
\caption{\textbf{\Method}: a workflow-aware benchmark for autonomous medical AI research. \textbf{Left:} Tasks are sourced from 20+ public challenges (e.g., KiTS19~\citep{kits19}) spanning diverse modalities (CT, MRI, X-ray, ultrasound, video) and task types (segmentation, detection, VQA, report generation, and image enhancement). Each task provides a natural-language description and deliverable target, with two difficulty tiers: \textit{Lite} (method, environment, and skill scaffolding provided) and \textit{Standard} (agent selects method and environment autonomously with plan-only guidance). \textbf{Right:} Given data access and a task description, an AI agent conducts auto research via a shared S1--S5 workflow --- Plan (understand task, select method), Setup (install dependencies, load models), Validate (run pilot case, inspect outputs, fix errors), Inference (run inference, write predictions), and Submit (verify schema, submit answers) --- before scoring and evaluation. Each agent operates in an isolated container with a private workspace; shared data and skill files are readable, but access to other agents' workspaces, evaluation ground truth, and scoring rubrics is prohibited, with violations triggering a warning then termination.}
\label{fig:workflow}
\end{figure}

\section{AutoMedBench}
\label{sec:method}

In this section, we introduce \Method, a workflow-aware benchmark for evaluating autonomous agents in end-to-end medical-AI research. Unlike static medical benchmarks that assess final predictions from fixed inputs~\citep{medqa2021,pubmedqa2019,medmcqa2022,medmnist2023,vqarad2018,pathvqa,slake}, \Method requires agents to complete a full research workflow: planning a solution, setting up the environment, validating the pipeline, performing inference, and submitting the required artifacts within a controlled research environment~\citep{agentbench2024,swebench2023,mlebench2024,paperbench2025}. The benchmark follows three design principles: (i) realistic medical research artifacts, (ii) process-level supervision of the research workflow, and (iii) deterministic held-out evaluation using private references.

\subsection{Benchmark Construction}

\paragraph{Task suite.}
\Method covers five medical research tracks, defined by the final artifact that the agent is required to produce: segmentation masks~\citep{ronneberger2015unet,isensee2021nnunet}, restored images~\citep{chen2017redcnn,zbontar2018fastmri}, VQA answers~\citep{vqarad2018,pathvqa,slake,lin2023medvqa}, reports~\citep{chen2020r2gen}, and detection boxes~\citep{ren2015fasterrcnn,redmon2016yolo,carion2020detr}. The 24 tasks span CT, MRI, X-ray, pathology, blood-smear microscopy, dental imaging, and medical video. We include a task only if its public inputs can be made available to the agent, its references can remain hidden from the agent while still enabling deterministic evaluation~\citep{mlebench2024,paperbench2025}, and its workflow can be expressed under the shared research protocol described below. Table~\ref{tab:tasks} enumerates the active tasks.

\begin{table}[t]
\centering
\scriptsize
\setlength{\tabcolsep}{4pt}
\renewcommand{\arraystretch}{0.95}
\newcommand{\release}[2]{\makebox[1.7em][r]{#1}\hspace{0.35em}#2}
\caption{Active tasks in \Method. We evaluate 24 tasks across five medical research tracks. Each task is assessed under two difficulty tiers, \textsc{Lite} and \textsc{Standard}, yielding \totaltasknum{} task-tier settings in total. For each track, the header reports the evaluation metric shared by all tasks in that track. The release column denotes the month of the first public dataset, challenge, or paper release, highlighting that \Method encompasses both well-established and recently introduced medical-AI benchmarks.}
\vspace{6pt}
\label{tab:tasks}
\begin{tabular}{@{}p{0.27\linewidth}p{0.26\linewidth}p{0.27\linewidth}p{0.12\linewidth}@{}}
\toprule
\textbf{Task} & \textbf{Dataset} & \textbf{Modality} & \textbf{Release} \\
\midrule

\rowcolor{trackgray}
\multicolumn{4}{@{}l@{}}{\rule{0pt}{2.1ex}\rule[-0.9ex]{0pt}{0pt}\textbf{Segmentation}~\textit{(macro Dice)}} \\
Kidney Tumor & KiTS19~\citep{kits19} & abdominal CT & \release{Mar}{2019} \\
Fetal Brain Tissues & FeTA~\citep{feta_challenge} & fetal MRI & \release{May}{2021} \\
Multi-Organ & TotalSegmentator~\citep{totalsegmentator} & whole-body CT & \release{Sep}{2023} \\
Airway Tree & AeroPath~\citep{aeropath} & thoracic CT & \release{Nov}{2023} \\
PANTHER T1 & PANTHER~\citep{panther_challenge} & T1-w MR-Linac & \release{Apr}{2025} \\
PANTHER T2 & PANTHER~\citep{panther_challenge} & T2-w MR-Linac & \release{Apr}{2025} \\
Pancreas Tumor & PanTS~\citep{pants_bodymaps} & abdominal CT & \release{Jul}{2025} \\
Pancreas OAR & PanTS~\citep{pants_bodymaps} & abdominal CT & \release{Jul}{2025} \\

\midrule
\rowcolor{trackgray}
\multicolumn{4}{@{}l@{}}{\rule{0pt}{2.1ex}\rule[-0.9ex]{0pt}{0pt}\textbf{Enhancement}~\textit{(SSIM)}} \\
LDCT Denoising & LDCT-SimNICT~\citep{ldct_aapm} & low-dose CT & \release{Jan}{2016} \\
MRI Super-Resolution & fastMRI~\citep{zbontar2018fastmri} & knee/brain MRI & \release{Nov}{2018} \\

\midrule
\rowcolor{trackgray}
\multicolumn{4}{@{}l@{}}{\rule{0pt}{2.1ex}\rule[-0.9ex]{0pt}{0pt}\textbf{VQA}~\textit{(accuracy)}} \\
Radiology VQA & VQA-RAD~\citep{vqarad2018} & radiology & \release{Nov}{2018} \\
Pathology VQA & PathVQA~\citep{pathvqa} & histopathology & \release{Mar}{2020} \\
Semantic Radiology VQA & SLAKE~\citep{slake} & radiology & \release{Feb}{2021} \\
Expert Multimodal VQA & MedXpertQA-MM~\citep{medxpertqa_mm} & mixed multimodal & \release{Jan}{2025} \\
Multi-frame Medical VQA & MedFrameQA~\citep{medframeqa} & medical video & \release{May}{2025} \\

\midrule
\rowcolor{trackgray}
\multicolumn{4}{@{}l@{}}{\rule{0pt}{2.1ex}\rule[-0.9ex]{0pt}{0pt}\textbf{Report Generation}~\textit{(report score)}} \\
Chest X-ray Findings/Impression & IU X-Ray~\citep{iu_xray} & chest X-ray & \release{Jul}{2015} \\
Chest X-ray Findings & MIMIC-CXR~\citep{mimic_cxr} & chest X-ray & \release{Aug}{2019} \\
Pathology Captioning 100 & PathCap~\citep{pathcap} & histopathology & \release{Mar}{2024} \\
Pathology Captioning 500 & PathCap~\citep{pathcap} & histopathology & \release{Mar}{2024} \\
Chest X-ray Full Report & CheXpert Plus~\citep{chexpert_plus} & chest X-ray & \release{May}{2024} \\

\midrule
\rowcolor{trackgray}
\multicolumn{4}{@{}l@{}}{\rule{0pt}{2.1ex}\rule[-0.9ex]{0pt}{0pt}\textbf{Detection}~\textit{(mAP@0.5)}} \\
Blood Cell & BCCD~\citep{bccd} & blood smear & \release{Dec}{2017} \\
Chest X-ray Abnormality & VinDr-CXR~\citep{vindr_cxr} & chest X-ray & \release{Jun}{2021} \\
Wrist Anomaly & GRAZPEDWRI-DX~\citep{grazpedwri_dx} & pediatric wrist X-ray & \release{May}{2022} \\
Dental Disease & DENTEX~\citep{dentex} & dental X-ray & \release{Apr}{2023} \\
\bottomrule
\end{tabular}
\end{table}

We exclude tasks whose correctness depends primarily on subjective judgment, long-horizon clinical dialogue~\citep{healthbench2025,schmidgall2024agentclinic,hicks2026healthbenchprofessionalevaluatinglarge}, or training-time adaptation~\citep{mlebench2024}. This design keeps the benchmark focused on inference-time medical-AI research workflows with stable and reproducible evaluation.

\paragraph{Shared workflow.}
Every task in \Method follows the same five-stage research workflow: \emph{Plan}, \emph{Setup}, \emph{Validate}, \emph{Inference}, and \emph{Submit}.
A run is defined as a continuous interaction between a base LLM agent and a code-execution environment.
The agent receives the task brief, public inputs, allowed public resources, and a writable workspace.
Held-out references are never visible to the agent and are mounted only inside the offline evaluator after the run terminates.

\newcolumntype{Y}{>{\RaggedRight\arraybackslash}X}
\newcommand{\stage}[2]{\makecell[c]{\textbf{#1}\\[-1pt]\footnotesize #2}}

\begin{table}[t]
\centering
\small
\setlength{\tabcolsep}{6pt}
\renewcommand{\arraystretch}{1.12}
\caption{The unified five-stage workflow adopted by all tasks in \Method.}
\vspace{4pt}
\label{tab:workflow-stages}
\begin{tabularx}{\linewidth}{@{}c c Y c@{}}
\toprule
\textbf{Stage} & \textbf{Skill} & \textbf{Required Work} & \textbf{Weight} \\
\midrule
\stage{S1}{\emph{Plan}} 
& Knowledge 
& Understand the task, select a feasible method, and write \texttt{plan.md}.
& \textbf{25\%} \\

\addlinespace[2pt]
\stage{S2}{\emph{Setup}} 
& Engineering 
& Install dependencies, load models or APIs, and verify paths and outputs. 
& \textbf{15\%} \\

\addlinespace[2pt]
\stage{S3}{\emph{Validate}} 
& Engineering 
& Run a pilot case, inspect intermediate outputs, and correct pipeline errors. 
& \textbf{35\%} \\

\addlinespace[2pt]
\stage{S4}{\emph{Inference}} 
& Engineering 
& Run full inference and generate prediction files.
& \textbf{15\%} \\

\addlinespace[2pt]
\stage{S5}{\emph{Submit}} 
& Engineering 
& Verify the submission schema and submit the final artifacts. 
& \textbf{10\%} \\
\bottomrule
\end{tabularx}
\end{table}

The workflow is designed to make the research process checkable, rather than evaluating only the final output~\citep{paperbench2025,bedi2026healthadminbench,liu2026physicianbench}.
To this end, each stage within the workflow requires supporting evidence either on disk or in the execution trace.
In particular, S1--S3 capture the main research decisions: selecting a method, preparing the environment, and validating the pipeline before scaling to the full task. S4--S5 capture execution completeness and submission validity. This shared workflow allows \Method to compare otherwise heterogeneous medical tasks under a common process-level protocol.

\paragraph{Post-run error coding.}
In addition to scoring workflow completion, \Method records cause-based error codes after each run for diagnostic analysis.
The detailed run report may contain multiple fired error codes, because a single run can show several error patterns during planning, setup, validation, execution, and submission.
Specifically, the benchmark harness saves the full interaction record as \texttt{conversation.json}, which is used to identify which error-code categories appear in the run.
The error codes are independent of the S1--S5 workflow stages: stage scores measure where the agent made progress in the required workflow, whereas fired error codes describe what types of breakdowns occurred.
We use five error-code categories: E1 understanding error, E2 data/model setup error, E3 verification or recovery error, E4 implementation or execution error, and E5 deliverable or submission error.
Clean successful runs receive no error code.
Error labels are used only for analysis and do not affect the \textsc{Agentic}, \textsc{Task}, or \textsc{Overall} scores.
Detailed definitions and examples are provided in Appendix~\ref{app:error-codes}.

\paragraph{Execution environment.}
Each run is conducted with a single base LLM serving as the agent.
We do not introduce vendor-specific agent frameworks, multi-agent controllers, or external retrieval wrappers.
All agents interact with tasks through the same code-execution interface, ensuring that performance differences primarily reflect model behavior under a fixed benchmark harness rather than task-specific orchestration.

Each task is executed under two-container isolation.
The agent container has GPU access, network access, a mounted public-input view, and a writable workspace.
In contrast, the offline evaluator container has access to the held-out references and scoring code, is isolated from external network communication, and receives only the submitted artifacts after the agent run terminates.
For each dataset, the benchmark harness materializes a public input view and a private reference store. The agent is granted read access only to the public view and write access only to its workspace, while private references are never mounted into the agent container.

To ensure fair and reproducible evaluation under this isolation design, \Method enforces an inference-only protocol.
Agents may use pre-trained models and approved model-inference APIs, but may not train or fine-tune models during a run.
If a run attempts to access private data, write outside the designated workspace, bypass the sandbox, or otherwise violate the isolation policy, the benchmark harness flags the run and assigns zero scores to all S1--S5 workflow stages.
The run is nevertheless retained in the cost ledger, ensuring that invalid attempts are included in resource analyses.

\begin{figure}[t]
  \centering
  \includegraphics[width=\textwidth]{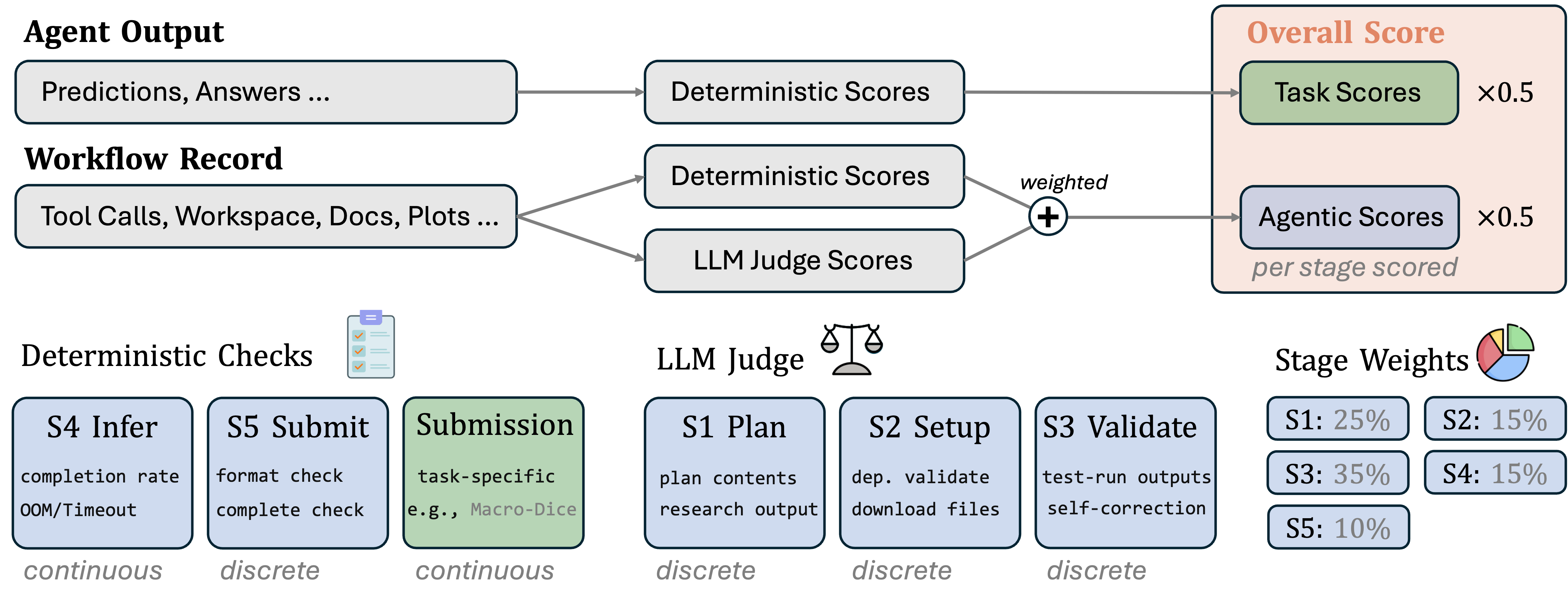}
\caption{\textbf{\Method scoring rubrics.} The overall score is the equal-weighted average of Task Score and Agentic Score ($\times 0.5$ each). \textbf{Task Score} is computed deterministically from agent predictions or answers. \textbf{Agentic Score} combines deterministic checks and LLM judge scores across the S1--S5 workflow stages, weighted as: S1 Plan (25\%), S2 Setup (15\%), S3 Validate (35\%), S4 Inference (15\%), and S5 Submit (10\%). S1, S2, and S3 are evaluated as discrete scores via LLM judge (plan contents, dependency validation, and self-correction); S4 is continuous (completion rate, OOM/timeout); S5 is discrete (format and completeness check). Task-specific metrics (e.g., Macro-Dice) are scored continuously and folded into the Task Score.}
  \label{fig:scoring_rubrics}
\end{figure}

\subsection{Task Formulation}

Given the benchmark construction above, we formulate each task instance in \Method as a unified research problem consisting of public inputs $\mathcal{D}_{pub}$, hidden references $\mathcal{D}_{priv}$, a task brief $b$, the final artifact $\mathcal{A}$ produced by the agent and evaluated by the evaluator, a submission schema $\mathcal{S}$, a task-specific metric $m$, and a wall-time limit $\tau$:
\[
\mathcal{T} = (\mathcal{D}_{pub}, \mathcal{D}_{priv}, b, \mathcal{A}, \mathcal{S}, m, \tau).
\]
Given $\mathcal{D}_{pub}$ and $b$, the agent must produce $\mathcal{A}$ such that it conforms to $\mathcal{S}$ within time $\tau$.
The evaluator then computes the task outcome as $m(\mathcal{A}, \mathcal{D}_{priv})$.

\paragraph{Agent objective.}
The agent is expected to produce a valid evaluated artifact through the full research workflow, rather than simply emit a final answer.
This requires not only selecting an appropriate method, but also making the pipeline executable, validating intermediate outputs, and submitting artifacts that conform to the required schema.
The formulation accommodates diverse solution strategies while preserving a fixed interface for reproducible evaluation.

\paragraph{Difficulty tiers.}
\Method instantiates two difficulty tiers, \textsc{Lite} and \textsc{Standard}, by varying only the amount of scaffolding provided in the task brief.
Across tiers, the input data, held-out references, wall-time limit, task metric, scoring code, and submission schema remain fixed.
Thus, the tiers control the degree of agent autonomy while keeping the underlying task unchanged.

\begin{itemize}[leftmargin=*]
    \item \textbf{\textsc{Lite}.} The brief identifies a viable method, specifies key dependencies, and provides stage-specific hints for planning, setup, and pilot validation. This tier evaluates whether an agent can execute a prescribed medical-AI workflow end to end.

    \item \textbf{\textsc{Standard}.} The brief specifies only bounded model or method families and leaves the final implementation unspecified. The agent must select an approach, resolve dependencies, and design validation checks independently. This tier evaluates whether an agent can make bounded methodological choices and implement them within the same end-to-end workflow.
\end{itemize}

Applying these two tiers to the 24 active tasks yields \totaltasknum{} task-tier settings evaluated in this paper.

\subsection{Evaluation Protocol}

We evaluate each run along two complementary axes: workflow execution and final artifact quality.
All component scores are computed in $[0,1]$ and reported as percentages, unless otherwise specified.
The top-line score is defined as:
\[
\textsc{Overall} = 0.5 \cdot \textsc{Agentic} + 0.5 \cdot \textsc{Task},
\]
where \textsc{Agentic} measures the agent's completion of the required research workflow, and \textsc{Task} measures the quality of the final artifact against held-out references.

\paragraph{Agentic workflow score.}
\textsc{Agentic} is a weighted sum of the five workflow stages:
\[
\textsc{Agentic} = 0.25\,\textsc{S1} + 0.15\,\textsc{S2} + 0.35\,\textsc{S3} + 0.15\,\textsc{S4} + 0.10\,\textsc{S5}.
\]
Each stage score lies in $[0,1]$, with different stages scored according to their evidence type. S1--S3 are evaluated as LLM judge scores from saved artifacts and execution traces~\citep{paperbench2025,bedi2026healthadminbench,liu2026physicianbench}, as they involve qualitative decisions in planning, setup, and validation. S4--S5 are evaluated through deterministic checks: S4 verifies that the expected prediction files exist for the evaluation inputs, and S5 verifies that the submitted artifacts conform to the required schema. The weights reflect the relative research consequence of each stage. S3 (Validate) receives the highest weight (35\%) because catching and correcting pipeline errors before full inference is the most critical and often neglected step in a research workflow. S1 (Plan) receives the second highest weight (25\%) because a flawed method choice or misunderstood task objective cannot be recovered downstream. S2, S4, and S5 receive lower weights as they are more mechanical: setting up dependencies, running inference, and submitting artifacts are necessary but less consequential than the core research decisions made in S1 and S3.

\paragraph{Task outcome score.}
\textsc{Task} is the standard held-out metric for each track, scaled to $[0,1]$. We use macro Dice~\citep{dice1945} for segmentation, mean SSIM~\citep{wang2004ssim} for image enhancement, VQA accuracy~\citep{antol2015vqa} for visual question answering, and mAP at IoU $0.5$~\citep{everingham2010voc} for detection. For report generation, we use the unweighted mean of BLEU~\citep{papineni2002bleu}, METEOR~\citep{banerjee2005meteor}, ROUGE-L~\citep{lin2004rouge}, F1RadGraph~\citep{jain2021radgraph}, and micro precision, recall, and F1. Invalid, missing, or unreadable outputs are handled according to the failure rules described below. Exact formulas are provided in Appendix~\ref{app:scoring}.

\paragraph{Failure handling.}

To maintain reproducible and policy-compliant evaluation, we define deterministic rules for incomplete, malformed, and invalid runs. When a run times out, the evaluator considers only artifacts written to the workspace before termination, and assigns zero to missing outputs for the corresponding cases. When a submission is malformed, S5 is set to zero, and the task metric is evaluated only if the submitted artifacts can be safely parsed. Runs that violate the isolation policy are marked invalid; all S1--S5 workflow stages are assigned zero, and the submitted artifacts are excluded from task scoring. This protocol preserves partial credit for valid intermediate progress while preventing malformed or policy-violating runs from receiving undue credit.

\subsection{Comparison with Existing Benchmarks}

\Method{} differs from existing medical and healthcare agent benchmarks by introducing a unique combination of medical auto-research challenges, as summarized in Table~\ref{tab:benchmark_comparison}. First, tasks in \Method{} require agents to complete end-to-end medical-AI research workflows and submit valid artifacts, rather than merely answer medical questions, conduct clinical dialogue, interact with EHR/FHIR systems, or operate healthcare administration portals. This setting requires agents to reason about the research objective, execute code, and produce outputs that can be evaluated by task-specific evaluators.
Second, \Method{} covers heterogeneous medical-AI tasks across five tracks: segmentation, image enhancement, VQA, report generation, and detection. In contrast to benchmarks centered on a single clinical environment or task family, \Method{} spans diverse modalities and artifact formats across radiology, pathology, microscopy, dental imaging, and medical video.
Third, \Method{} evaluates both the research process and the final outcome. Whereas many prior benchmarks primarily report final answer accuracy or task success, \Method{} adopts a shared cross-task workflow with explicit workflow-level scoring. Finally, \Method{} enables more diagnostic evaluation through hidden or post-execution checks, controlled difficulty tiers, and post-run error diagnosis.
\begin{table}[t]
\centering
\scriptsize
\setlength{\tabcolsep}{3.2pt}
\renewcommand{\arraystretch}{0.95}
\caption{\textbf{Comparison with medical and healthcare agent benchmarks.}
``Full Med-AI Pipeline'' denotes end-to-end medical-AI pipeline evaluation.
``Code Env.'' denotes access to a code-execution research environment.
``Cross-task Workflow'' denotes a unified workflow shared across tasks.
``Workflow Score'' denotes process-level or checkpoint-based scoring.
``Hidden Eval.'' denotes hidden references, held-out states, blind tests, or post-execution checks unseen by the agent.
``Tiers'' denotes controlled difficulty levels.
``Error Diag.'' denotes post-run error diagnosis.}
\vspace{4pt}
\label{tab:benchmark_comparison}
\resizebox{\linewidth}{!}{%
\begin{tabular}{@{}lccccccc@{}}
\toprule
\textbf{Benchmark} &
\textbf{Full Med-AI Pipeline} &
\textbf{Code Env.} &
\textbf{Cross-task Workflow} &
\textbf{Workflow Score} &
\textbf{Hidden Eval.} &
\textbf{Tiers} &
\textbf{Error Diag.} \\
\midrule

HealthBench Professional~\citep{hicks2026healthbenchprofessionalevaluatinglarge}
& \xmark & \xmark & \xmark & \xmark & \xmark & \xmark & \xmark \\

MedHELM~\citep{bedi2025medhelm}
& \xmark & \xmark & \xmark & \xmark & \cmark & \xmark & \xmark \\

AgentClinic~\citep{schmidgall2024agentclinic}
& \xmark & \xmark & \xmark & \xmark & \cmark & \xmark & \xmark \\

EHRAgent~\citep{shi2024ehragent}
& \xmark & \cmark & \xmark & \xmark & \xmark & \cmark & \xmark \\

MedAgentBench~\citep{jiang2025medagentbenchrealisticvirtualehr}
& \xmark & \xmark & \cmark & \xmark & \cmark & \cmark & \xmark \\

FHIR-AgentBench~\citep{lee2025fhir}
& \xmark & \cmark & \xmark & \cmark & \xmark & \xmark & \xmark \\

AgentEHR~\citep{liao2026agentehr}
& \xmark & \xmark & \cmark & \xmark & \xmark & \xmark & \cmark \\

HealthAdminBench~\citep{bedi2026healthadminbench}
& \xmark & \xmark & \cmark & \cmark & \cmark & \cmark & \xmark \\

PhysicianBench~\citep{liu2026physicianbench}
& \xmark & \xmark & \cmark & \cmark & \cmark & \xmark & \cmark \\

CamylaBench~\citep{gao2026camyla}
& \cmark & \cmark & \cmark & \xmark & \cmark & \xmark & \cmark \\

\midrule
\textbf{\Method{}}
& \cmark & \cmark & \cmark & \cmark & \cmark & \cmark & \cmark \\

\bottomrule
\end{tabular}
}
\end{table}

\begin{figure}[t]
    \centering
    \includegraphics[width=\linewidth]{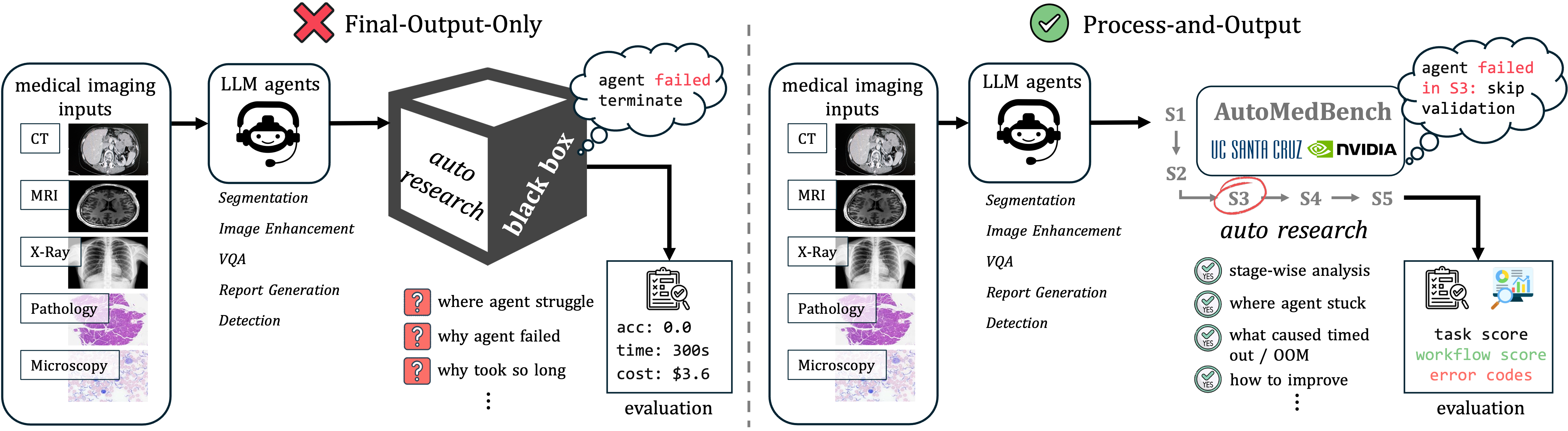}
    \caption{\textbf{\Method provides stage-level evaluation for medical research agents.}
    Unlike most prior benchmarks that only measure the final output, \Method tracks the full workflow from planning to submission, making it possible to identify where agents fail during the research process. This process-aware evaluation reveals hidden failure modes, workflow weaknesses, and error patterns that are not visible from final task scores alone.}
    \label{fig:teaser_figure}
\end{figure}

\section{Experimental Setup}
\label{sec:exp}

We evaluate \Method under a fixed agent interface and a unified replication protocol, following the controlled evaluation practice used in execution-grounded agent benchmarks~\citep{agentbench2024,mlebench2024,paperbench2025}. This section describes the agents, task-tier coverage, replication protocol, and logging procedures used in the analyses in Section~\ref{sec:results}.

\subsection{Agents}

We evaluate six frontier base models on the \totaltasknum{} \textsc{Lite}/\textsc{Standard} task-tier settings in \Method, covering both hosted proprietary models and open-weight models served through our own inference stack. Appendix~\ref{app:model-details} lists the model names, vendors, release dates, and open-source status. Each model is used directly as the agent in a single long-horizon interaction with the same code-execution environment. To isolate the effect of the underlying base model, we keep the system prompt, tool schema, stop conditions, and default decoding settings fixed across models. We do not add vendor-specific agent wrappers, multi-agent controllers, or task-specific retrieval pipelines beyond the shared benchmark interface, since scaffold and orchestration choices can materially affect agent results~\citep{mlebench2024,paperbench2025}. As a consequence, the comparison reflects differences among frontier base models under the same benchmark contract rather than differences among product-level agent systems.

\subsection{Evaluation Runs and Logging}

\paragraph{Task coverage and data access.}
The main evaluation covers all 24 active tasks across five medical research tracks under both \textsc{Lite} and \textsc{Standard}, yielding \totaltasknum{} reported task-tier settings. All tasks follow the public-input/private-reference split described in Section~\ref{sec:method}. \Method does not redistribute restricted datasets; runners must obtain any credentialed data before launching the benchmark harness. In our experiments, MIMIC-CXR is accessed through PhysioNet~\citep{mimic_cxr}, and fastMRI is accessed under the NYU data-sharing agreement~\citep{zbontar2018fastmri}. When an official evaluation script is available from the source benchmark, we execute it inside the offline evaluation container rather than re-implementing the metric.

\paragraph{Replication protocol.}
The smallest evaluation unit is an \emph{evaluation cell}, defined by an \mbox{(agent, task, tier)} tuple. With six agents, 24 tasks, and two tiers, the main experiments comprise 288 evaluation cells. Each replicate starts from the same task-specific container image and a fresh writable workspace, with no shared cache, files, or conversation history across runs. The default cohort size is $N{=}10$ runs per cell. Five segmentation tasks---KiTS19, PanTS Tumor, PanTS OAR, FeTA, and AeroPath---use $N{=}20$ runs to better estimate performance under longer execution horizons and higher observed run-to-run variance. Each task has a fixed wall-time cap that is held constant across agents and difficulty tiers. A run terminates when the agent submits successfully or when the wall-time cap is reached; upon timeout, the evaluator scores only artifacts already written to the workspace.

\paragraph{Logging and cost accounting.}
For every run, we log the five workflow stage scores, the derived \textsc{Task} and \textsc{Overall} scores, the number of conversational turns, wall-clock time, input and output token counts, estimated inference cost, run status, and the full interaction record in \texttt{conversation.json}. Post-run diagnostics are derived from this interaction record: the detailed report records all fired error codes from E1--E5, following the rubric in Appendix~\ref{app:error-codes}. Each run writes one row to a unified ledger keyed by $(\textit{agent}, \textit{task}, \textit{tier}, \textit{run-id})$, allowing all reported statistics to be recomputed without replaying the benchmark. For cost accounting, we normalize all costs using the fixed rate snapshot in Table~\ref{tab:appendix-price-snapshot}, without prompt-cache or negotiated discounts, following the resource-accounting emphasis in execution-heavy agent benchmarks~\citep{mlebench2024,paperbench2025}. Resource records and error-code labels are used only for analysis and do not affect workflow or task scoring.
We define an end-to-end completed run as a run that submits artifacts accepted by the evaluation module and receives a task score.
A failed run is one that does not reach this end-to-end state.
For runs with two or more fired error codes, we define recovery as still reaching end-to-end completion after those errors appear in the detailed report.
Accordingly, recovery rate is the percentage of runs with at least two fired error codes that still submit scoreable artifacts.

\section{Results and Analysis}
\label{sec:results}

We report three levels of evidence. Section~\ref{sec:main_results} presents the overall leaderboard and summarizes where current agents stand on \Method. Section~\ref{sec:diagnostic_analysis} uses workflow, tier, and cost analyses to diagnose when and why agents struggle. Section~\ref{sec:failure_analysis} examines fine-grained failure modes and recovery behavior.

\subsection{Main Results}
\label{sec:main_results}

\begin{table}[ht]
\centering
\scriptsize
\setlength{\tabcolsep}{2pt}
\renewcommand{\arraystretch}{0.95}
\caption{\textbf{Per-track and overall leaderboard.} Scores are averaged over all runs for the tasks and tiers within each track. Agent rows are ordered by the overall leaderboard rank, the overall column is shown on the right, and the highest score in each column is highlighted.}
\vspace{6pt}
\label{tab:overall_leaderboard}
\resizebox{\linewidth}{!}{%
\begin{tabular}{@{}lcccccc@{}}
\toprule
\textbf{Agent} & \textbf{Segmentation} & \textbf{Enhancement} & \textbf{Visual Question Answering} & \textbf{Report Generation} & \textbf{Detection} & \textbf{Overall} \\
\midrule
\textbf{Opus 4.6~\citep{anthropic_opus46}} & \cellcolor{bestgreen}67.2 & \cellcolor{bestgreen}78.3 & 55.5 & \cellcolor{bestgreen}55.8 & \cellcolor{bestgreen}85.7 & \cellcolor{bestgreen}66.5 \\
\textbf{GLM-5~\citep{zhipu_glm5}} & 58.0 & 68.8 & \cellcolor{bestgreen}64.0 & 48.6 & 83.3 & 61.6 \\
\textbf{Gemini 3.1 Pro~\citep{google_gemini31_pro}} & 54.7 & 70.9 & 62.3 & 47.7 & 77.2 & 59.0 \\
\textbf{ChatGPT-5.4~\citep{openai_gpt54}} & 59.4 & 75.1 & 36.5 & 40.8 & 73.8 & 55.3 \\
\textbf{MiniMax-M2.5~\citep{minimax_m25}} & 46.5 & 74.5 & 55.8 & 28.9 & 80.0 & 51.6 \\
\textbf{Qwen3.5~\citep{alibaba_qwen35}} & 42.8 & 63.5 & 57.0 & 38.7 & 81.4 & 51.2 \\
\bottomrule
\end{tabular}
}
\end{table}

\begin{figure}[t]
    \centering
    \includegraphics[width=\linewidth]{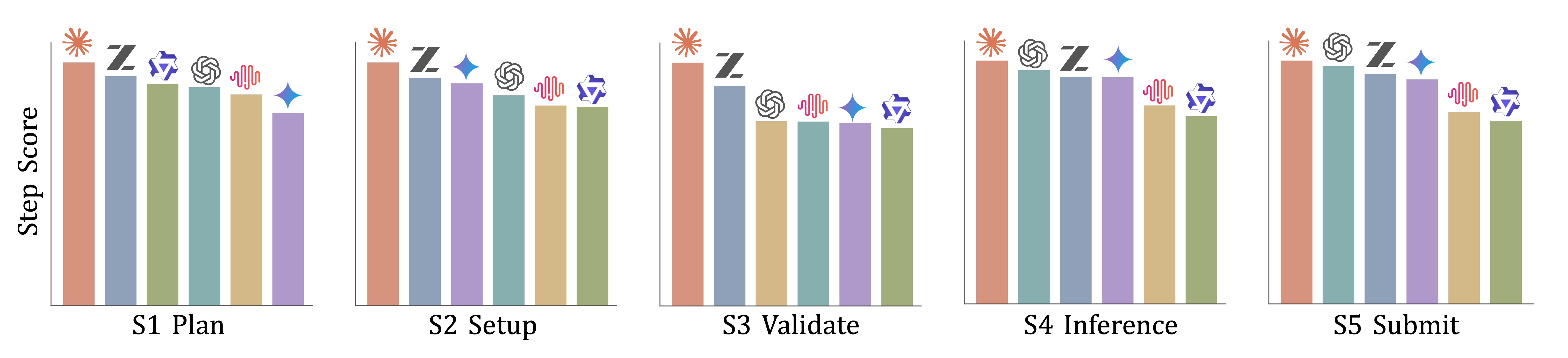}
    \caption{\textbf{Step-level workflow scoring across agents.} Scores are shown for the six evaluated agents at each workflow stage: S1 Plan, S2 Setup, S3 Validate, S4 Inference, and S5 Submit. The dashed line marks the mean score across agents for each stage. Setup is the strongest stage on average, while validation is the weakest, showing that agents are better at making pipelines runnable than at checking whether those pipelines are reliable before full inference and submission. A strong agent tends to perform consistently well across steps, as seen for Opus 4.6, whereas other agents show more uneven profiles, such as GPT-5.4.}
    \label{fig:step_scoring}
\end{figure}

\paragraph{Finding 1: The leaderboard separates agents but does not identify a uniformly best profile.}
Figure~\ref{fig:leaderboard} and Table~\ref{tab:overall_leaderboard} show a 15.3-point spread in overall score across the evaluated agents, from 51.2 to 66.5. The top overall agent also leads segmentation, enhancement, report generation, and detection, while a different agent leads visual question answering. The step-level breakdown in Figure~\ref{fig:step_scoring} further shows that agents with similar overall scores can differ in where they succeed or fail within the workflow. This pattern suggests that \Method captures meaningful differences between agents while also exposing track-specific and process-specific strengths and weaknesses. We therefore treat the leaderboard as a starting point for diagnosis rather than as a single-number measure of medical research ability.

\paragraph{Finding 2: Task quality lags behind workflow completion.}
A consistent gap appears between agentic and task scores: all evaluated agents score higher on the workflow component than on the final task component. This suggests that agents are often able to make progress through the required stages, but the resulting medical artifacts remain substantially weaker than their apparent workflow progress would imply. In other words, completing the visible steps of an auto-research process does not guarantee that the final segmentation mask, restored image, VQA answer, report, or detection output is correct.

\paragraph{Finding 3: Medical tracks expose different agent weaknesses.}
The per-track results in Table~\ref{tab:overall_leaderboard} show that performance varies strongly across medical research tracks. Detection obtains the highest scores for several agents, suggesting that constrained output formats and mature pretrained detectors make these tasks comparatively easier under our benchmark~\citep{ren2015fasterrcnn,redmon2016yolo,carion2020detr}. Report generation and VQA are more challenging, likely because they require semantic interpretation of medical images and text beyond producing a valid artifact~\citep{lin2023medvqa,chen2020r2gen}. Segmentation remains competitive for the best agents but is costly and pipeline-heavy, especially for 3D medical volumes~\citep{litjens2017survey,isensee2021nnunet}. These differences indicate that no single track is sufficient to characterize medical auto-research ability; agents can appear strong in one artifact type while failing in another.

\subsection{Diagnostic Analysis}
\label{sec:diagnostic_analysis}

\paragraph{Validation is the central workflow bottleneck.}
Figure~\ref{fig:step_scoring} breaks agent performance down by the five workflow stages. S3 (\emph{Validate}) has the lowest mean score across agents, while S2 (\emph{Setup}) is the highest. This pattern suggests that agents are better at installing dependencies and preparing a runnable environment than at designing and executing meaningful pilot checks before scaling to full inference. The stage-level view also reveals differences hidden by a single final score: agents with similar overall performance can differ in where they fail, such as late-stage inference and submission versus early planning and validation. This supports the need for workflow-aware scoring rather than final-output evaluation alone, consistent with recent rubric- or checkpoint-based agent benchmarks~\citep{paperbench2025,bedi2026healthadminbench,liu2026physicianbench}.

\begin{table}[t]
\centering
\scriptsize
\setlength{\tabcolsep}{4pt}
\renewcommand{\arraystretch}{1.05}
\caption{\textbf{More scaffolding does not consistently improve agentic scores.}
\textsc{Lite} and \textsc{Standard} use the same data, metric, time cap, scoring code, and submission schema, but \textsc{Lite} provides more detailed scaffolding. $\Delta$ reports the relative change from \textsc{Standard} to \textsc{Lite}, computed as $(\textsc{Lite}-\textsc{Standard})/\textsc{Standard}\times100$. Green values indicate improvement under \textsc{Lite}; red values indicate a drop. See tier details in \tableautorefname~\ref{tab:appendix-tier-details}.}
\vspace{6pt}
\label{tab:tier_compare}
\resizebox{\linewidth}{!}{%
\begin{tabular}{@{}lcccccc@{}}
\toprule
 & \textbf{Opus 4.6} & \textbf{GLM-5} & \textbf{Gemini 3.1 Pro} & \textbf{GPT-5.4} & \textbf{MiniMax-M2.5} & \textbf{Qwen3.5-397B} \\
\midrule
\textsc{Standard}
& 81.8 & 71.9 & 69.7 & 78.9 & 65.3 & 61.7 \\
\textsc{Lite}
& 81.1 & 77.9 & 71.7 & 66.0 & 66.4 & 66.6 \\
\midrule
$\Delta$
& \textcolor{red!70!black}{\textbf{-0.9}}
& \textcolor{green!50!black}{\textbf{+8.3}}
& \textcolor{green!50!black}{\textbf{+2.8}}
& \textcolor{red!70!black}{\textbf{-16.3}}
& \textcolor{green!50!black}{\textbf{+1.7}}
& \textcolor{green!50!black}{\textbf{+8.0}} \\
\bottomrule
\end{tabular}
}
\end{table}

\paragraph{More scaffolding does not always improve performance.}
Table~\ref{tab:tier_compare} compares \textsc{Standard} and \textsc{Lite} agentic scores across agents. Since the underlying data, metric, time limit, scoring code, and submission schema are fixed across tiers, the only change is the amount of task-brief scaffolding. Moving from \textsc{Standard} to the more detailed \textsc{Lite} tier does not produce a clear, uniform improvement: four agents improve under \textsc{Lite}, but two agents perform worse, including a 16.3\% relative drop for GPT-5.4. This suggests that additional scaffolding is not automatically beneficial; in some cases it may constrain the agent to a brittle workflow, encourage unnecessary steps, or induce inefficient behavior. This sensitivity is important because prior agent benchmarks often evaluate full model--scaffold systems rather than base models alone~\citep{mlebench2024,paperbench2025}.

\begin{figure}[t]
    \centering
    \includegraphics[width=\linewidth]{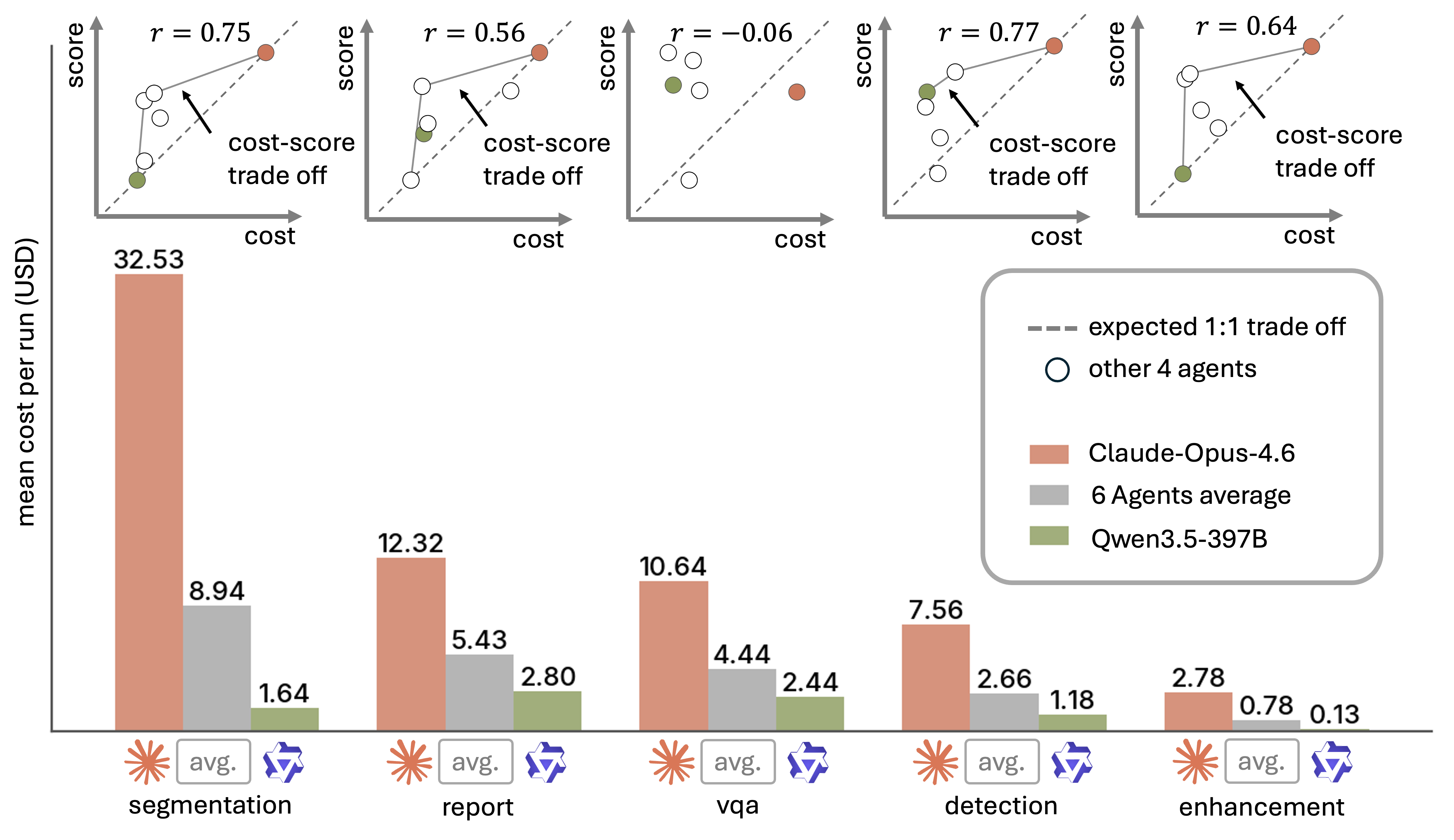}
    \caption{\textbf{Higher cost does not reliably translate into better performance.} Bars show the mean cost per run for each task track. Insets plot each agent's track-level cost against score, with Pearson correlation $r$ summarizing the cost--performance relationship within that track. The weak and track-dependent correlations indicate that raw spending is not the main driver of success. The API-price snapshot is listed in Table~\ref{tab:appendix-price-snapshot}.}
    \label{fig:cost_analysis}
\end{figure}

\paragraph{Higher cost does not reliably translate into better performance.}
Figure~\ref{fig:cost_analysis} shows that \Method spans a wide range of cost regimes, from relatively cheap enhancement tasks to much more expensive segmentation tasks. Within each track, higher spending is only weakly associated with better performance. Segmentation shows the clearest positive cost--performance relationship, but report generation, detection, and enhancement show diminishing returns, and VQA shows almost no relationship. In many cases, the gap in cost between agents is larger than the gap in score. These results suggest that raw spending is not the main driver of success; what matters more is whether agents use compute effectively for validation, debugging, and recovery. Resource-aware reporting is therefore important for execution-heavy agent evaluation~\citep{mlebench2024,paperbench2025}.

\paragraph{Absolute performance and cost efficiency identify different agents.}
The leaderboard and cost analysis point to different deployment choices. Table~\ref{tab:overall_leaderboard} shows that Opus 4.6 obtains the highest overall score, while the resource summary in Appendix~\ref{app:resource-statistics} shows that it also has the highest average cost per run. By contrast, GLM-5 reaches the second-best overall score at lower average cost. This suggests that the highest-scoring agent and a lower-cost practical choice need not be the same. For repeated benchmark runs across datasets, tiers, and task variants, \Method therefore supports both capability-oriented comparison and resource-aware model selection.

\begin{table}[t]
\centering
\scriptsize
\setlength{\tabcolsep}{4pt}
\renewcommand{\arraystretch}{1.12}
\caption{\textbf{Cause-based error codes.} A run may fire multiple codes when multiple breakdown types appear in the trace.}
\vspace{4pt}
\label{tab:error-code-summary}
\resizebox{\linewidth}{!}{%
\begin{tabular}{@{}ccccc@{}}
\toprule
\textbf{E1 Understanding} & \textbf{E2 Setup} & \textbf{E3 Verification} & \textbf{E4 Execution} & \textbf{E5 Submission} \\
\midrule
\makecell[c]{hallucination\\wrong model type}
& \makecell[c]{wrong dependency\\failed model load}
& \makecell[c]{skipped validation\\ignored bad output}
& \makecell[c]{runtime crash\\shape mismatch}
& \makecell[c]{missing files\\wrong format} \\
\bottomrule
\end{tabular}
}
\end{table}

\begin{figure}[t]
    \centering
    \includegraphics[width=\linewidth]{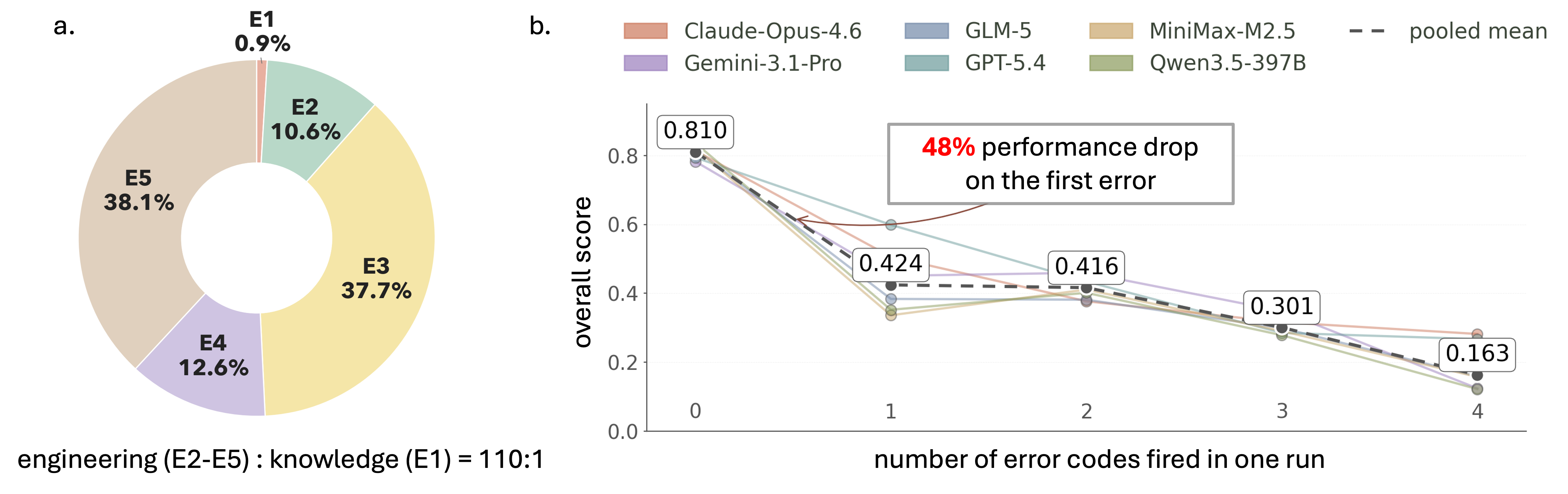}
    \caption{\textbf{Error codes can sharply derail a run.}
    (a) Distribution of fired error-code types.
    (b) Mean overall score by the number of fired error codes in a run. Verification and submission errors dominate tagged failures. The first fired error produces a large score drop, and runs with two or more fired errors remain in a low-score regime.}
    \label{fig:error_statistics}
\end{figure}

\subsection{Fine-Grained Failure Analysis}
\label{sec:failure_analysis}

After each run, the detailed report records all fired error codes observed in the trace. We analyze these fired codes directly to summarize which breakdown types occur most often. Table~\ref{tab:error-code-summary} summarizes the five codes used in the following analysis.

\paragraph{Engineering-shaped failures dominate agent breakdowns.}
Figure~\ref{fig:error_statistics} summarizes the post-run error-code categories defined in Appendix~\ref{app:error-codes}. Panel (a) reports percentages over all fired code tags. Most tagged errors in \Method are engineering-shaped rather than understanding-shaped. Setup, execution, verification, and deliverable errors account for the majority of observed failures, while E1 understanding errors are rare. This does not imply that agents have sufficient medical knowledge; instead, it shows that in our end-to-end setting, many runs fail through practical research-workflow problems: invalid environments, failed execution, missed validation signals, incomplete outputs, missing files, or malformed submissions. Similar engineering and execution bottlenecks have been observed in software- and ML-engineering agent benchmarks~\citep{swebench2023,mlebench2024,paperbench2025}.

\begin{figure}[htbp]
    \centering
    \includegraphics[width=\linewidth]{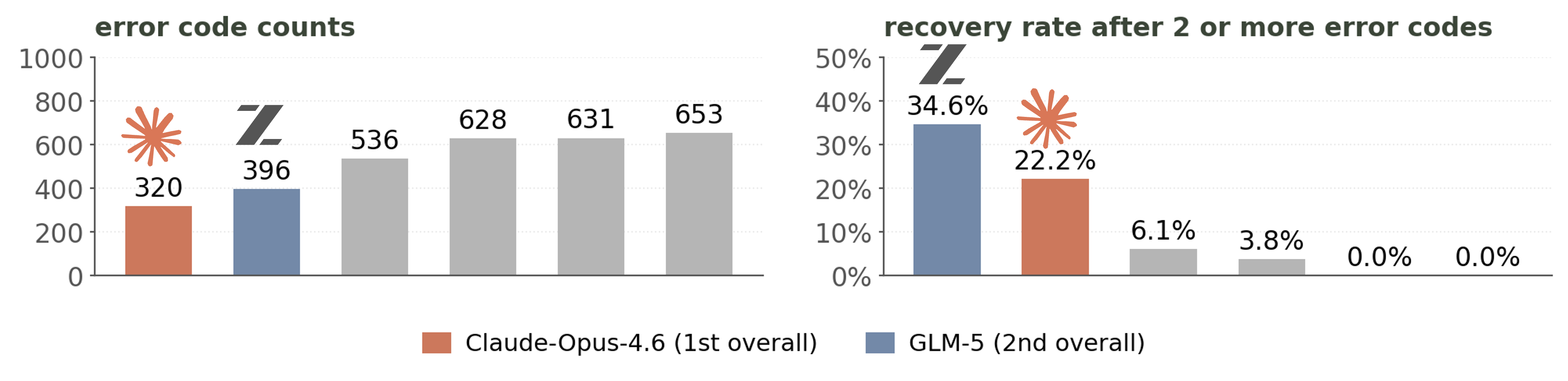}
    \caption{\textbf{Strong agents both avoid errors and recover from them.} Left: total fired error-code counts across agents. Right: recovery rate after two or more fired error codes, defined as the percentage of such runs that still reach end-to-end completion and receive an evaluation score.}
    \label{fig:what_makes_a_good_agent}
\end{figure}

\paragraph{Fired error codes mark a performance cliff.}
Figure~\ref{fig:error_statistics}b shows that runs with fired error codes score much lower than runs with no fired errors. On average, runs with one fired error code have a 48\% lower overall score than runs with no fired error code, and runs with two or more fired codes remain in a low-score regime. This motivates the recovery analysis in Figure~\ref{fig:what_makes_a_good_agent}, where recovery is measured as reaching end-to-end completion after at least two fired error codes. These results highlight the importance of early error detection and recovery in medical auto-research agents.

\paragraph{Strong agents are better at recovering after multiple fired errors.}
Figure~\ref{fig:what_makes_a_good_agent} shows that top-performing agents are not simply the ones with the fewest fired error codes. Instead, stronger agents more often recover after two or more errors appear in the interaction record. This finding refines the interpretation of agent reliability: success depends not only on avoiding early mistakes, but also on persisting through debugging when the workflow begins to fail. Error avoidance alone is insufficient, and persistence without a clean recovery strategy is also insufficient. Robust medical auto-research agents need both a stable start and the ability to repair the workflow after multiple failures occur.

\paragraph{The largest improvement opportunity is workflow control, not only model knowledge.}
The step-level and error-code analyses point to a common pattern. Figure~\ref{fig:step_scoring} shows that agents are comparatively strong at setup but weaker at validation, while Figure~\ref{fig:error_statistics} and Table~\ref{tab:error-code-definitions} show that many failures arise from execution, verification, and deliverable errors rather than from E1 understanding errors. This suggests that improving medical research agents requires more than stronger task reasoning or larger base models. Agents need explicit validation routines, artifact-level sanity checks, and recovery policies that connect observed failures to corrective actions before full inference or submission.

\section{Related Work}
\label{sec:related}

\paragraph{Agentic Evaluation.}
Recent agent benchmarks move beyond static question answering by placing models in interactive environments with external state, tools, and verifiable artifacts. AgentBench, AgentBoard, and GAIA test broad multi-step reasoning; WebArena, AssistantBench, and OSWorld ground agents in browser or operating-system environments; SWE-bench, Terminal-Bench, MLE-bench, and PaperBench emphasize execution-heavy tasks with sandboxed or repository-based grading~\citep{agentbench2024,agentboard2024,gaia2024,webarena2024,assistantbench2024,osworld2024,swebench2023,terminalbench2026,mlebench2024,paperbench2025}. These works motivate \Method's environment-grounded evaluation, but remain largely domain-general and do not model medical data constraints, modality-specific metrics, or staged medical research workflows.

\paragraph{Research Automation.}
Research-agent work studies whether LLM agents can assist scientific discovery by writing code, analyzing data, running experiments, or reproducing research artifacts~\citep{mlagentbench2024,discoveryworld2024,aiscientist2024,corebench2025,paperbench2025}. This direction is closest to \Method because it evaluates agents as research workers rather than passive answer generators. However, most settings focus on general scientific workflows~\citep{discoveryworld2024,aiscientist2024}, machine-learning engineering~\citep{mlagentbench2024,mlgym2025,mlebench2024}, or paper replication and reproducibility~\citep{corebench2025,paperbench2025}. \Method targets medical-AI research specifically, where agents must handle medical data separation, modality-dependent outputs, and a shared Plan--Setup--Validate--Inference--Submit workflow.

\paragraph{Medical Benchmarks.}
Medical AI benchmarks typically evaluate fixed-input knowledge, clinical reasoning, or single-task prediction. Language-centric datasets such as MedQA, PubMedQA, MedMCQA, and MultiMedQA test exam-style or biomedical question answering, while HealthBench and MedHELM move toward richer rubric-based evaluation in realistic health scenarios~\citep{medqa2021,pubmedqa2019,medmcqa2022,multimedqa2023,healthbench2025,medhelm2026}.
Medical imaging and multimodal benchmarks extend evaluation to visual tasks, including MedMNIST v2 for lightweight biomedical image classification and VQA-RAD, PathVQA, and SLAKE for image-grounded medical question answering~\citep{medmnist2023,vqarad2018,pathvqa,slake}. MedAgentBench is closest in spirit, but targets clinical EHR assistance rather than medical model building~\citep{jiang2025medagentbenchrealisticvirtualehr}. In contrast, \Method evaluates whether agents can set up environments, validate pipelines, run inference, recover from workflow failures, and submit valid artifacts under medical metrics.

\section{Conclusion}
\label{sec: conclution}

We present \Method, a workflow-aware benchmark for evaluating autonomous agents on end-to-end medical-AI research tasks. Unlike prior benchmarks that measure only final output quality, \Method evaluates both how agents work and what they produce, using a shared five-stage workflow across 24 tasks and five medical research tracks. Our evaluation protocol combines process-level agentic scoring, deterministic held-out task metrics, controlled difficulty tiers, and post-run error diagnosis, enabling a more complete picture of where and why agents succeed or fail in medical auto-research.

Our experiments with six frontier models show that current agents remain far from reliable medical-AI researchers. While agents can often set up runnable pipelines, validation is consistently the weakest stage, and engineering failures dominate over understanding errors. These findings suggest that the main bottleneck is not medical knowledge alone, but the ability to verify intermediate outputs and recover from workflow errors. We hope \Method provides a practical foundation for building agents that can conduct medical-AI research more reliably and systematically.

\bibliographystyle{iclr2026_conference}
\bibliography{references}

\appendix
\clearpage
\section*{Appendix Contents}

\newcommand{\appendixcontentsline}[2]{%
    \noindent #1\dotfill \pageref{#2}\par
}
\newcommand{\appendixcontentssubline}[2]{%
    \noindent\hspace{1.5em}#1\dotfill \pageref{#2}\par
}

\appendixcontentsline{Per-Task Scoring Definitions}{app:scoring}
\appendixcontentsline{Scoring Rubrics Details}{app:scoring-rubrics-details}
\appendixcontentsline{Evaluated Model Details}{app:model-details}
\appendixcontentsline{Run Resource Statistics}{app:resource-statistics}
\appendixcontentssubline{API Price Snapshot}{app:api-price-snapshot}
\appendixcontentsline{Difficulty Tier Details}{app:tier-details}
\appendixcontentsline{Workflow Step Details}{app:workflow-step-details}
\appendixcontentsline{Error-Code Definitions}{app:error-codes}
\appendixcontentsline{Per-Task Scoring Details}{app:per-task-scoring-details}
\appendixcontentsline{Example Benchmarking Traces}{app:example-benchmarking-traces}

\clearpage
\section{Per-Task Scoring Definitions}
\label{app:scoring}

This appendix gives the exact per-task metric used in $\textsc{Task}$ for each track. Every metric is scaled to $[0,1]$ and averaged over the $N$ cases in the held-out set; missing or unreadable outputs receive 0 for the affected case.

\paragraph{Segmentation: macro Dice~\citep{dice1945}.}
For case $i$ with $K_i$ targets, prediction $P_{ik}$, and reference mask $G_{ik}$, we use the standard Dice overlap $2|P_{ik}\cap G_{ik}|\,/\,(|P_{ik}|+|G_{ik}|)$, averaged first over the $K_i$ targets in a case and then over cases.

\paragraph{Enhancement: mean SSIM~\citep{wang2004ssim}.}
For restored image $\hat{x}_i$ and private reference $x_i$, the case score is $\operatorname{SSIM}(\hat{x}_i, x_i)$, averaged over cases.

\paragraph{VQA: exact-match accuracy~\citep{antol2015vqa}.}
With normalized prediction $\hat{a}_i$ and gold answer $a_i$, the case score is $\mathbf{1}\{\hat{a}_i = a_i\}$, averaged over cases.

\paragraph{Reports: averaged text and entity metrics.}
Each case gets BLEU~\citep{papineni2002bleu}, METEOR~\citep{banerjee2005meteor}, ROUGE-L~\citep{lin2004rouge}, F1RadGraph~\citep{jain2021radgraph}, and micro precision, recall, and F1. The case score $s_i$ is the unweighted mean of these seven values; the task score is the mean of $s_i$ over cases.

\paragraph{Detection: mAP@0.5~\citep{everingham2010voc}.}
We follow the PASCAL VOC protocol and report mean average precision at IoU $0.5$, averaged over the $C$ classes in the task.

\clearpage
\section{Scoring Rubrics Details}
\label{app:scoring-rubrics-details}

Table~\ref{tab:appendix-segmentation-stage-rubric} gives segmentation as one concrete example of the workflow-scoring rubric. This example follows the public segmentation evaluator.\footnote{\url{https://github.com/AutoMedBench/AutoMedBench/tree/eval_seg/eval_seg}} For other task tracks, please check the public GitHub repository.\footnote{\url{https://github.com/AutoMedBench/AutoMedBench}}

\begin{table}[htbp]
\centering
\scriptsize
\setlength{\tabcolsep}{4pt}
\renewcommand{\arraystretch}{1.12}
\caption{\textbf{Segmentation workflow-scoring rubric details.} S1--S3 use LLM judge scores from saved artifacts and execution traces. S4--S5 use deterministic checks from the evaluator.}
\vspace{6pt}
\label{tab:appendix-segmentation-stage-rubric}
\begin{tabularx}{\linewidth}{@{}p{0.11\linewidth}p{0.08\linewidth}p{0.14\linewidth}X p{0.09\linewidth}@{}}
\toprule
\textbf{Step} & \textbf{Item} & \textbf{Score type} & \textbf{Segmentation rubric} & \textbf{Value} \\
\midrule
S1 Plan & S1a & LLM judge score & \texttt{plan.md} exists. & $\{0,1\}$ \\
 & S1b & LLM judge score & \texttt{plan.md} gives clear pipeline instructions; scored 0 if \texttt{plan.md} is missing. & $\{0,1\}$ \\
 & S1c & LLM judge score & The selected model covers the lesion/tumor target or all required tissue labels and mappings; scored 0 if \texttt{plan.md} is missing. & $\{0,1\}$ \\
 & S1d & LLM judge score & At least three distinct models are researched. In Lite, this item is credited because the model is given. & $\{0,1\}$ \\
 & S1e & LLM judge score & \texttt{plan.png} exists. In Lite, this item is credited because the plot is not required. & $\{0,1\}$ \\
 & S1f & LLM judge score & The plot shows a clear pipeline diagram; scored 0 if the plot is missing. In Lite, this item is credited. & $\{0,1\}$ \\
 & S1 score & Average & $(\mathrm{S1a}+\mathrm{S1b}+\mathrm{S1c}+\mathrm{S1d}+\mathrm{S1e}+\mathrm{S1f})/6$. & $[0,1]$ \\
\midrule
S2 Setup & S2a & LLM judge score & Model checkpoint or weights are successfully downloaded. & $\{0,1\}$ \\
 & S2b & LLM judge score & Input compatibility is checked, including spacing, shape, or dtype when relevant. & $\{0,1\}$ \\
 & S2c & LLM judge score & Environment setup succeeds, including virtual environment or package installation. & $\{0,1\}$ \\
 & S2d & LLM judge score & Environment failures are resolved within five attempts, or no such failure occurs. & $\{0,1\}$ \\
 & S2e & LLM judge score & The model is loaded on GPU and confirmed working. & $\{0,1\}$ \\
 & S2 score & Average & $(\mathrm{S2a}+\mathrm{S2b}+\mathrm{S2c}+\mathrm{S2d}+\mathrm{S2e})/5$. & $[0,1]$ \\
\midrule
S3 Validate & S3=1.0 & LLM judge score & A pilot patient is tested before batch inference and output shape/values are checked. For multi-class segmentation, allowed labels and per-tissue coverage are checked. & $1.0$ \\
 & S3=0.5 & LLM judge score & Some validation is performed, but it is incomplete, such as checking shape without checking lesion or tissue coverage. & $0.5$ \\
 & S3=0.0 & LLM judge score & No validation is detected, or the agent runs batch inference immediately without verifying outputs. & $0.0$ \\
\midrule
S4 Inference & S4a & Deterministic check & Completion rate: fraction of expected patients with output files. & $[0,1]$ \\
 & S4b & Deterministic check & Mask-format validity: all masks are readable and have valid values and the expected shape. & $\{0,1\}$ \\
 & S4 score & Formula & $0.5 \times \mathrm{S4a} + 0.5 \times \mathrm{S4b}$. & $[0,1]$ \\
\midrule
S5 Submit & S5a & Deterministic check & Valid-results flag: at least one patient is scored and has positive Dice evidence. & $\{0,1\}$ \\
 & S5b & Deterministic check & Output-format validity: all masks pass the evaluator format check. & $\{0,1\}$ \\
 & S5 score & Formula & $0.5 \times \mathrm{S5a} + 0.5 \times \mathrm{S5b}$. & $\{0,0.5,1\}$ \\
\bottomrule
\end{tabularx}
\end{table}

For segmentation task scoring, incomplete runs receive zero Dice credit if any expected patient output is missing. S4 still records the partial completion rate for workflow diagnosis.

\clearpage
\section{Evaluated Model Details}
\label{app:model-details}

Table~\ref{tab:models} lists the six base models used in the main experiments. The set includes hosted proprietary models and open-weight models so that \Method measures agent performance across both API-served and self-hosted deployment modes. All models are evaluated through the same benchmark harness, with the same prompt template, tool schema, workspace layout, stopping rules, and scoring scripts. The ``open-source'' column indicates whether the model weights are publicly available; it does not change the task interface or the scoring procedure.

\begin{table}[h]
\centering
\small
\caption{Base models evaluated in this paper.}
\label{tab:models}
\setlength{\tabcolsep}{4pt}
\begin{tabular}{@{}p{0.32\linewidth}p{0.28\linewidth}p{0.18\linewidth}c@{}}
\toprule
\textbf{Name} & \textbf{Vendor} & \textbf{Release date} & \textbf{Open-source} \\
\midrule
\texttt{chatgpt-5.4}        & OpenAI~\citep{openai_gpt54}          & Mar. 5, 2026 & No \\
\texttt{gemini-3.1-pro}     & Google DeepMind~\citep{google_gemini31_pro} & Feb. 19, 2026 & No \\
\texttt{qwen3.5}            & Alibaba~\citep{alibaba_qwen35}       & Feb. 16, 2026 & Yes \\
\texttt{minimax-m2.5}       & MiniMax~\citep{minimax_m25}          & Feb. 12, 2026 & Yes \\
\texttt{glm-5}              & Zhipu AI / THU~\citep{zhipu_glm5}    & Feb. 11, 2026 & Yes \\
\texttt{claude-opus-4.6}    & Anthropic~\citep{anthropic_opus46}   & Feb. 5, 2026 & No \\
\bottomrule
\end{tabular}
\end{table}

\clearpage
\section{Run Resource Statistics}
\label{app:resource-statistics}

Table~\ref{tab:appendix-resource-stats} reports average resource use per run for the six evaluated agents. The averages are computed from the public leaderboard run summaries, exclude Kimi, and are weighted by the number of runs in each task-tier setting. Agent order follows the overall leaderboard ranking used in the main text.

The average overall cost per run is \$19.77 for Opus 4.6, \$2.73 for GLM-5, \$5.85 for Gemini 3.1 Pro, \$3.94 for ChatGPT-5.4, \$2.70 for MiniMax-M2.5, and \$1.83 for Qwen3.5.

\begin{table}[htbp]
\centering
\small
\setlength{\tabcolsep}{5pt}
\renewcommand{\arraystretch}{1.0}
\begin{threeparttable}
\caption{\textbf{Average resource use per run.} Time is wall-clock minutes, turns are conversational turns, tokens are total LLM tokens, and cost is normalized USD under the rate snapshot described in \S\ref{sec:exp}.}
\vspace{6pt}
\label{tab:appendix-resource-stats}
\begin{tabular}{@{}lcccc@{}}
\toprule
\textbf{Agent} & \textbf{Avg. time (min)} & \textbf{Avg. turns} & \textbf{Avg. tokens} & \textbf{Avg. cost} \\
\midrule
Opus 4.6 & 28.9 & 35.0 & 1.27M & \$19.77 \\
GLM-5 & 30.8 & 45.9 & 1.34M & \$2.73 \\
Gemini 3.1 Pro & 23.3 & 27.3 & 0.82M & \$5.85 \\
ChatGPT-5.4 & 22.7 & 14.0 & 0.25M & \$3.94 \\
MiniMax-M2.5 & 30.3 & 43.2 & 1.34M & \$2.70 \\
Qwen3.5 & 29.4 & 31.5 & 0.88M & \$1.83 \\
\bottomrule
\end{tabular}
\begin{tablenotes}[flushleft]
\footnotesize
\item \textit{Note:} Average cost is computed from platform-reported run charges, not by multiplying total tokens by text-only token rates; the price snapshot is for reference only.
\end{tablenotes}
\end{threeparttable}
\end{table}

Table~\ref{tab:appendix-resource-by-track} reports the same resource fields averaged by task track. These per-track values summarize the task-level settings that feed the cost analysis in Figure~\ref{fig:cost_analysis}.

\begin{table}[htbp]
\centering
\small
\setlength{\tabcolsep}{5pt}
\renewcommand{\arraystretch}{1.0}
\caption{\textbf{Average resource use per run by task track.} Values are averaged over agents, tiers, and task settings within each track, excluding Kimi and weighted by run count.}
\vspace{6pt}
\label{tab:appendix-resource-by-track}
\begin{tabular}{@{}lcccc@{}}
\toprule
\textbf{Task track} & \textbf{Avg. time (min)} & \textbf{Avg. turns} & \textbf{Avg. tokens} & \textbf{Avg. cost} \\
\midrule
Segmentation & 41.6 & 41.8 & 1.24M & \$8.98 \\
Enhancement & 27.9 & 24.7 & 0.40M & \$0.81 \\
VQA & 24.5 & 26.5 & 0.91M & \$4.44 \\
Report & 12.3 & 25.0 & 0.97M & \$5.43 \\
Detection & 4.9 & 24.9 & 0.52M & \$2.66 \\
\bottomrule
\end{tabular}
\end{table}

\subsection{API Price Snapshot}
\label{app:api-price-snapshot}

Table~\ref{tab:appendix-price-snapshot} lists the OpenRouter prices used for cost accounting. Prices are in USD per million tokens and were queried from the OpenRouter model API on May 28, 2026.

\begin{table}[htbp]
\centering
\scriptsize
\setlength{\tabcolsep}{4pt}
\renewcommand{\arraystretch}{1.05}
\caption{\textbf{API price snapshot.} Input and output prices are USD per million tokens. We apply these fixed rates to all runs, with no prompt-cache discounts or negotiated discounts. Model IDs follow OpenRouter.}
\vspace{6pt}
\label{tab:appendix-price-snapshot}
\begin{tabular}{@{}l>{\raggedright\arraybackslash}p{0.42\linewidth}cc@{}}
\toprule
\textbf{Agent} & \textbf{OpenRouter model ID} & \textbf{Input price} & \textbf{Output price} \\
\midrule
Opus 4.6 & \texttt{anthropic/claude-opus-4.6} & \$5.00 & \$25.00 \\
GLM-5 & \texttt{z-ai/glm-5} & \$0.72 & \$2.30 \\
Gemini 3.1 Pro & \texttt{google/gemini-3.1-pro-preview} & \$2.00 & \$12.00 \\
ChatGPT-5.4 & \texttt{openai/gpt-5.4} & \$2.50 & \$15.00 \\
MiniMax-M2.5 & \texttt{minimax/minimax-m2.5} & \$0.118 & \$0.99 \\
Qwen3.5 & \texttt{qwen/qwen3.5-397b-a17b} & \$0.39 & \$2.34 \\
\bottomrule
\end{tabular}
\end{table}

\clearpage
\section{Difficulty Tier Details}
\label{app:tier-details}

Tables~\ref{tab:appendix-tier-summary} and~\ref{tab:appendix-tier-details} summarize the difference between Lite and Standard. The two tiers use the same input data, held-out references, time limits, metrics, scoring code, and submission schema; only the task brief changes.

\begin{table}[htbp]
\centering
\scriptsize
\setlength{\tabcolsep}{6pt}
\renewcommand{\arraystretch}{1.05}
\caption{\textbf{Lite versus Standard at a glance.} Short side-by-side comparison of the task-brief scaffolding in each tier.}
\vspace{6pt}
\label{tab:appendix-tier-summary}
\begin{tabular}{@{}lcc@{}}
\toprule
\textbf{Dimension} & \textbf{Lite} & \textbf{Standard} \\
\midrule
Goal & Known workflow & Chosen workflow \\
Method & Concrete & Bounded \\
Dependencies & Pinned & Agent-resolved \\
Planning & Translate & Compare + justify \\
Setup & Recreate & Resolve \\
Validation & Guided & Self-designed \\
Research burden & Low & Moderate \\
Measured ability & Execution & Bounded choice \\
\bottomrule
\end{tabular}
\end{table}

\begin{table}[htbp]
\centering
\scriptsize
\setlength{\tabcolsep}{5pt}
\renewcommand{\arraystretch}{1.12}
\caption{\textbf{Detailed Lite versus Standard specification.} Both tiers keep the same data, references, metrics, time limits, scoring code, submission schema, and S1--S5 workflow; the table expands what changes in the brief.}
\vspace{6pt}
\label{tab:appendix-tier-details}
\begin{tabular}{@{}p{0.22\linewidth}p{0.34\linewidth}p{0.34\linewidth}@{}}
\toprule
\textbf{Dimension} & \textbf{Lite} & \textbf{Standard} \\
\midrule
Goal & Execute a viable workflow end to end with the main method already specified. & Choose and execute a suitable workflow within bounded method families. \\
\midrule
Method guidance & Names a concrete method or model family that is known to work for the task. & Gives candidate families or constraints, but leaves the final method choice to the agent. \\
\midrule
Dependency guidance & Pins key packages, scripts, checkpoints, or APIs when these are needed for a stable run. & Mentions required capabilities, but the agent must identify compatible packages, checkpoints, or APIs. \\
\midrule
Planning expectation & Translate the supplied workflow into \texttt{plan.md}. & Compare plausible approaches and justify the selected workflow in \texttt{plan.md}. \\
\midrule
Setup expectation & Recreate the provided environment recipe and verify that the named components run. & Resolve environment choices, install compatible dependencies, and verify that the selected components run. \\
\midrule
Validation support & Provides stage-specific hints for pilot validation, expected output shapes, and common failure modes. & Requires the agent to design its own pilot validation and decide what outputs are plausible. \\
\midrule
Research burden & Most research decisions are already scaffolded. & Method selection, dependency resolution, and validation design are part of the task. \\
\midrule
Primary measurement & Measures whether the agent can reliably execute a known medical-AI workflow. & Measures whether the agent can make bounded research choices and still complete the same workflow. \\
\bottomrule
\end{tabular}
\end{table}

Table~\ref{tab:appendix-workflow-stage-details} expands the compact workflow table in \S\ref{sec:method}. Each row names the concrete work expected from the agent and the artifact or check used by the harness. Consistent with Figure~\ref{fig:scoring_rubrics}, S1--S3 use LLM judge scores and S4--S5 use deterministic checks.

\clearpage
\section{Workflow Step Details}
\label{app:workflow-step-details}
\begin{table}[htbp]
\centering
\scriptsize
\setlength{\tabcolsep}{4pt}
\renewcommand{\arraystretch}{1.05}
\caption{\textbf{Detailed workflow-stage requirements.} The main text gives the short version; this table lists the corresponding expected work and evidence.}
\vspace{6pt}
\label{tab:appendix-workflow-stage-details}
\begin{tabular}{@{}p{0.13\linewidth}p{0.35\linewidth}p{0.42\linewidth}@{}}
\toprule
\textbf{Step} & \textbf{Detailed work} & \textbf{Evidence used for scoring} \\
\midrule
\multirow{3}{*}{S1 Plan} & Understand the task brief, target artifact, input files, output format, and task metric. & Notes in \texttt{plan.md} and consistency with the task brief. \\
 & Research feasible methods and select an approach that fits the task constraints. & Method choice and rationale in \texttt{plan.md}. \\
 & Write \texttt{plan.md} with execution steps, expected outputs, and validation checks. & Completed plan artifact saved in the workspace. \\
\midrule
\multirow{3}{*}{S2 Setup} & Install dependencies and prepare the software environment. & Successful commands, installed packages, and runnable scripts. \\
 & Load allowed pre-trained weights or configure allowed model-inference APIs. & Model/API availability in the execution trace. \\
 & Verify required data paths, scripts, and output directories. & Workspace files and setup checks before validation. \\
\midrule
\multirow{3}{*}{S3 Validate} & Run a pilot case or small public subset before full inference. & Pilot outputs and validation logs. \\
 & Inspect intermediate outputs for shape, format, and clinical plausibility. & Explicit validation notes or checks in the trace. \\
 & Fix setup or pipeline errors before scaling. & Evidence of debugging and corrected reruns. \\
\midrule
\multirow{2}{*}{S4 Inference} & Run the selected pipeline on the full evaluation input set. & Completed inference commands and generated outputs. \\
 & Write required prediction files for every evaluation case. & Output completeness checked by the harness. \\
\midrule
\multirow{2}{*}{S5 Submit} & Verify that saved predictions match the required submission schema. & Schema check or equivalent format validation. \\
 & Submit only final artifacts to the evaluator. & Submitted files passed to the offline evaluator. \\
\bottomrule
\end{tabular}
\end{table}

\clearpage
\section{Error-Code Definitions}
\label{app:error-codes}
\newcommand{\errcause}[1]{\textcolor{red}{#1}}

This appendix defines the post-run error codes used in the failure analysis. After the agent interaction ends, the detailed report records all fired error codes observed in the trace. The input is the recorded \texttt{conversation.json}, which contains the task prompt, agent messages, tool calls, command outputs, and submitted-file history. A run may fire multiple codes; these labels describe observed breakdown types, not a single exclusive cause. Error-code labels are diagnostic only and are not used to compute \textsc{Agentic}, \textsc{Task}, or \textsc{Overall}.

\begin{table}[htbp]
\centering
\scriptsize
\setlength{\tabcolsep}{4pt}
\renewcommand{\arraystretch}{1.1}
\caption{\textbf{Error-code rubric.} A run may fire multiple codes when multiple breakdown types appear in the trace.}
\vspace{6pt}
\label{tab:error-code-definitions}
\begin{tabular}{@{}p{0.08\linewidth}p{0.20\linewidth}p{0.47\linewidth}p{0.17\linewidth}@{}}
\toprule
\textbf{Code} & \textbf{Name} & \textbf{Definition} & \textbf{Common evidence} \\
\midrule
E1 & Understanding error & The agent solves the wrong problem or chooses a high-level approach incompatible with the task objective, modality, metric, constraints, or required artifact. & Incorrect task interpretation; incompatible method; hallucinated requirement. \\
\midrule
E2 & Data/model setup error & The agent understands the task but cannot correctly access, prepare, load, or configure required data, models, APIs, dependencies, or runtime resources. & Wrong paths; dependency conflicts; failed checkpoint/API/GPU loading. \\
\midrule
E3 & Verification/recovery error & The run produces evidence of invalid intermediate or final outputs, but the agent fails to detect, validate, debug, or recover from the problem. & Skipped sanity checks; ignored logs; accepted empty or implausible outputs. \\
\midrule
E4 & Implementation/execution error & The intended pipeline is plausible, but the agent's code, commands, or processing logic fail during execution. & Runtime exceptions; shape/type bugs; preprocessing bugs; partial execution. \\
\midrule
E5 & Deliverable/submission error & Usable outputs exist or could have been produced, but the final artifacts are missing, incomplete, malformed, wrongly named, misplaced, or incompatible with the evaluator schema. & Missing required files; wrong JSON/CSV/NIfTI format; incomplete case coverage. \\
\bottomrule
\end{tabular}
\end{table}

The detailed report applies these categories according to the observed evidence. E1 fires when the run solves the wrong problem or uses an incompatible high-level approach. E2 fires when the main blocker is preparing the data, model, dependencies, API, or runtime resources. E3 fires when warning signs or invalid outputs appear but are not detected or repaired. E4 fires when the intended pipeline fails while processing inputs. E5 fires when the main remaining failure is packaging or submitting the final artifacts.

\begin{table}[htbp]
\centering
\scriptsize
\setlength{\tabcolsep}{3pt}
\renewcommand{\arraystretch}{1.12}
\caption{\textbf{Examples of error codes by task track.} The examples illustrate how the same error-code taxonomy applies across heterogeneous medical artifacts.}
\vspace{6pt}
\label{tab:error-code-examples}
\resizebox{\linewidth}{!}{%
\begin{tabular}{@{}p{0.05\linewidth}p{0.19\linewidth}p{0.19\linewidth}p{0.19\linewidth}p{0.19\linewidth}p{0.19\linewidth}@{}}
\toprule
\textbf{Code} & \textbf{Segmentation} & \textbf{Enhancement} & \textbf{VQA} & \textbf{Report generation} & \textbf{Detection} \\
\midrule
E1 & Treats a mask-generation task as \errcause{image classification}, or chooses a method that cannot output voxel masks. & Treats MRI super-resolution as \errcause{denoising}, or optimizes for the wrong target resolution. & Answers \errcause{disease presence} when the task requires exact short-answer VQA. & Generates \errcause{captions} when the task requires structured radiology findings. & Uses \errcause{image-level classification} when bounding boxes are required. \\
\midrule
E2 & Cannot load \errcause{CT volumes, affine metadata, or a segmentation checkpoint}. & Fails to install \errcause{restoration dependencies} or load the pretrained denoising model. & Cannot load the \errcause{vision-language model, tokenizer, image files, or API key}. & Cannot access the \errcause{report model, sentence tokenizer, or image/report metadata}. & Fails to load \errcause{detector weights, class maps, or image annotation metadata}. \\
\midrule
E3 & Pilot masks are \errcause{empty or misaligned}, but the agent does not inspect or correct them. & Restored images are \errcause{blank, clipped, or unchanged}, but the agent accepts them. & Answers are \errcause{all identical or invalid}, but the agent skips sanity checks. & Reports are \errcause{repetitive, empty, or clinically implausible}, but the agent does not revise. & Boxes are \errcause{outside image bounds} or all confidence scores are zero, but the agent proceeds. \\
\midrule
E4 & Crashes from \errcause{tensor-shape mismatch}, wrong voxel orientation handling, or sliding-window inference bugs. & Produces runtime errors in \errcause{patch stitching, normalization, or image resizing}. & Breaks \errcause{batching or prompt construction}, causing empty or malformed answers. & Crashes while \errcause{decoding reports, parsing studies, or aligning generated text with cases}. & Crashes during \errcause{preprocessing, non-maximum suppression, or box coordinate conversion}. \\
\midrule
E5 & Masks are generated but saved with \errcause{wrong filenames, spacing, or NIfTI layout}. & Restored images exist but are submitted with \errcause{wrong extension, size, or directory layout}. & Answers exist but \errcause{JSON/CSV fields, case IDs, or normalization are wrong}. & Reports exist but \errcause{missing required study IDs, sections, or schema fields}. & Detections exist but boxes use the \errcause{wrong coordinate convention, labels, or file format}. \\
\bottomrule
\end{tabular}
}
\end{table}

\clearpage
\section{Per-Task Scoring Details}
\label{app:per-task-scoring-details}
\begin{figure}[htbp]
    \centering
    \includegraphics[width=\linewidth]{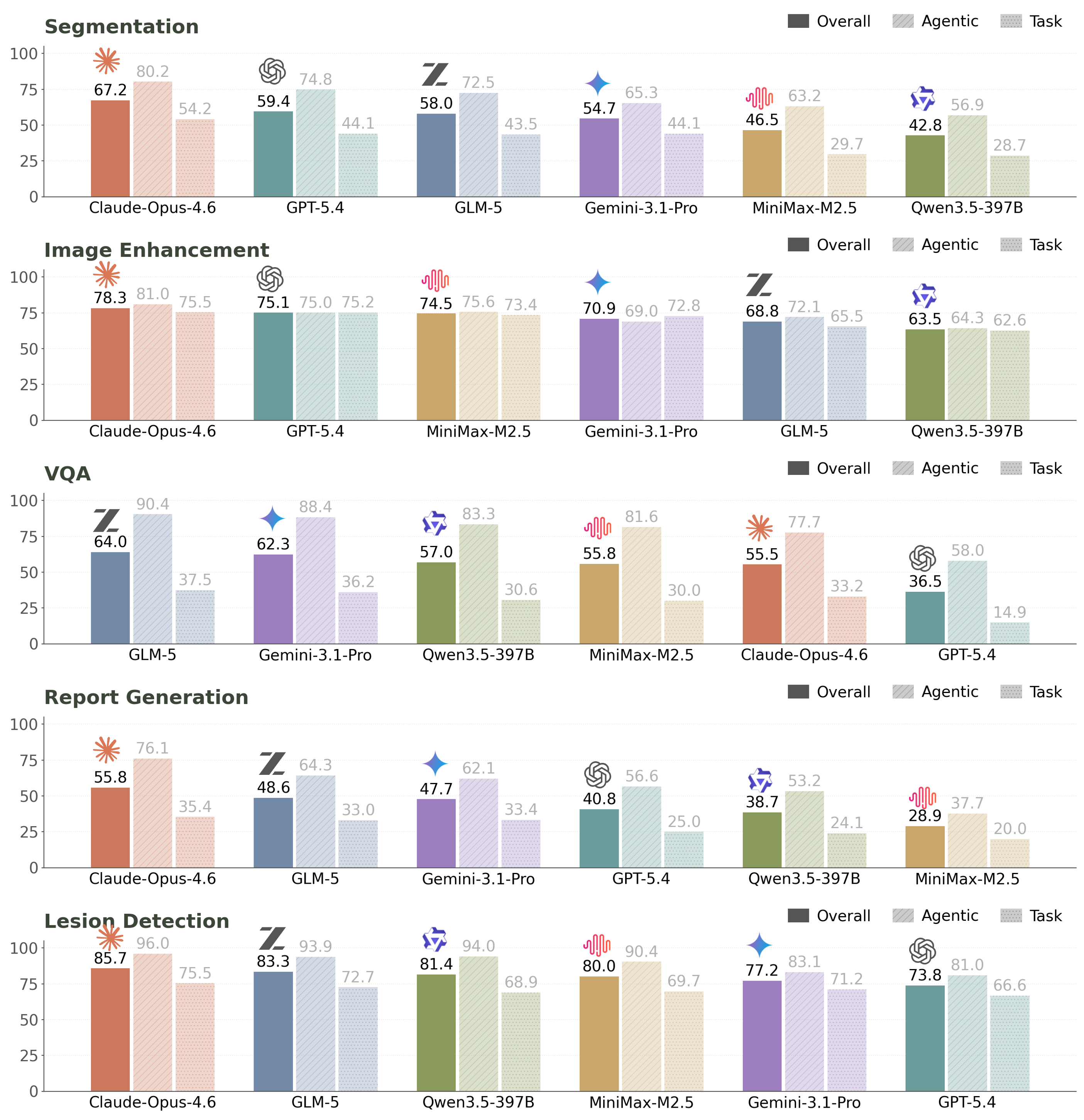}
    \caption{\textbf{Track-wise leaderboard breakdown.} Overall, agentic, and task scores are shown for each evaluated agent across segmentation, image enhancement, VQA, report generation, and lesion detection. The breakdown shows that overall rank masks task-track specialization: Opus 4.6 leads most tracks, while GLM-5 leads VQA and several agents remain competitive on detection.}
    \label{fig:task_wise_leaderboard}
\end{figure}

\clearpage
\section{Example Benchmarking Traces}
\label{app:example-benchmarking-traces}

This section shows two real kidney tumor segmentation runs. For each tier, we show the task text, the run report, API usage, and one short conversation example. Local paths, backend names, and secret-like strings are redacted.

\newcommand{\traceblocktitle}[1]{\vspace{0.9em}\noindent\textbf{#1}\par\vspace{0.3em}}

\subsection{Lite}
\label{app:example-benchmarking-traces-lite}
\traceblocktitle{Task description.}
\begin{lstlisting}[style=appendixtrace]
Task: kidney tumor segmentation
Tier: Lite
Input: 20 public CT cases.
Output: organ.nii.gz and lesion.nii.gz for each case.
Guidance: the model is given; the agent researches it and follows S1-S5.
Goal: make valid masks and submit them for held-out scoring.
\end{lstlisting}

\traceblocktitle{Detailed report.}
\begin{lstlisting}[style=appendixtrace]
Agent: Claude Opus 4.6
Tier: Lite
Task: kidney tumor segmentation

Scores:
  Agentic score: 0.7567
  Task score: 0.3073
  Overall score: 0.5320

Task metrics:
  Organ Dice: 0.9160
  Lesion Dice: 0.1045

Workflow scores:
  S1 Plan: 1.0000
  S2 Setup: 1.0000
  S3 Validate: 0.5000
  S4 Inference: 1.0000
  S5 Submit: 0.3166

Run cost:
  Estimated cost: $28.1496

Output checks:
  Submission valid: true
  Mask files valid: true
  CSV valid: false

Dominant error:
  E3 verification/recovery error
\end{lstlisting}

\traceblocktitle{API calls.}
\begin{lstlisting}[style=appendixtrace]
API calls: 55
Code runs: 54
Submit calls: 1
Input tokens: 1706325
Output tokens: 34063
Total tokens: 1740388
Estimated cost: $28.1496
Wall time: 1673.42 s

Phase summary:
  S1 Plan: 14 calls, 0 errors, 99.0 s
  S2 Setup: 3 calls, 1 error, 99.4 s
  S3 Validate: 9 calls, 2 errors, 118.3 s
  S4 Inference: 27 calls, 3 errors, 718.1 s
  S5 Submit: 2 calls, 0 errors, 71.3 s
\end{lstlisting}

\traceblocktitle{Example conversation.}
\begin{lstlisting}[style=appendixtrace]
USER:
Begin. The model architecture has been chosen for you. Research it, then follow S1 through S5.

ASSISTANT:
I'll start by researching MONAI VISTA3D and setting up the environment.

TOOL CALL:
execute_code: ls <PUBLIC_DATA>/

TOOL OUTPUT:
patient_00000001 ... patient_00000020

ASSISTANT:
I inspected one CT case, checked the VISTA3D labels, and wrote a plan.
The plan uses right kidney label 5, left kidney label 14, and kidney mass label 129.

TOOL OUTPUT:
CUDA available: True
GPU: NVIDIA H100 80GB HBM3
Model loaded successfully on cuda:0
Parameters: 218.0M

ASSISTANT:
I ran a pilot case before full inference.

TOOL OUTPUT:
Preprocessed image shape: [1, 313, 227, 204]
Output shape: [1, 3, 313, 227, 204]
Label 5 voxels: 57564
Label 14 voxels: 48060
Label 129 voxels: 7

ASSISTANT:
The lesion result is small, but the pipeline runs. I will run all cases and verify the files.

TOOL OUTPUT:
Checking 20 patients...
patient_00000001: OK
patient_00000002: OK
patient_00000003: OK

ASSISTANT:
All masks exist. I will submit the results.
\end{lstlisting}

\end{document}